\documentclass{article}

\PassOptionsToPackage{numbers, compress, square, sort}{natbib}

\usepackage[preprint]{neurips_2023}

\usepackage[utf8]{inputenc} 
\usepackage[T1]{fontenc}    
\usepackage{hyperref}       
\usepackage{url}            
\usepackage{booktabs}       
\usepackage{amsfonts}       
\usepackage{nicefrac}       
\usepackage{microtype}      
\usepackage{xcolor}         

\definecolor{tyellow1}{HTML}{FCE94F}
\definecolor{tyellow2}{HTML}{EDD400}
\definecolor{tyellow3}{HTML}{C4A000}
\definecolor{torange1}{HTML}{FCAF3E}
\definecolor{torange2}{HTML}{F57900}
\definecolor{torange3}{HTML}{C35C00}
\definecolor{tbrown1}{HTML}{E9B96E}
\definecolor{tbrown2}{HTML}{C17D11}
\definecolor{tbrown3}{HTML}{8F5902}
\definecolor{tgreen1}{HTML}{8AE234}
\definecolor{tgreen2}{HTML}{73D216}
\definecolor{tgreen3}{HTML}{4E9A06}
\definecolor{tblue1}{HTML}{729FCF}
\definecolor{tblue2}{HTML}{3465A4}
\definecolor{tblue3}{HTML}{204A87}
\definecolor{tpurple1}{HTML}{AD7FA8}
\definecolor{tpurple2}{HTML}{75507B}
\definecolor{tpurple3}{HTML}{5C3566}
\definecolor{tred1}{HTML}{EF2929}
\definecolor{tred2}{HTML}{CC0000}
\definecolor{tred3}{HTML}{A40000}
\definecolor{tlgray1}{HTML}{EEEEEC}
\definecolor{tlgray2}{HTML}{D3D7CF}
\definecolor{tlgray3}{HTML}{BABDB6}
\definecolor{tdgray1}{HTML}{888A85}
\definecolor{tdgray2}{HTML}{555753}
\definecolor{tdgray3}{HTML}{2E3436}

\usepackage{amsmath}
\usepackage{amssymb}
\usepackage{amsthm}
\usepackage{thmtools}
\usepackage{algorithm}
\usepackage[noEnd,indLines,commentColor=tblue2]{algpseudocodex}
\usepackage{float}
\usepackage{newfloat}
\usepackage{subcaption}
\usepackage{tikz}
\usetikzlibrary{arrows,decorations,fit,positioning,shapes}
\usepackage{empheq,fancybox}
\usepackage{adjustbox}
\usepackage{appendix}
\usepackage{stmaryrd}
\usepackage{multirow}
\usepackage{bm}

\graphicspath{{plots/},{plots/28,28_mnist_1v3_flip},{plots/28,28_mnist_1v7_flip},{plots/10,10_mnist_1v3_flip},{plots/10,10_mnist_1v7_flip}}

\DeclareGraphicsExtensions{.pdf}

\title{On Formal Feature Attribution and Its Approximation}

\newcommand{\mailtodomain}[1]{\href{mailto:#1@monash.edu}{\nolinkurl{#1}}}

\author{%
  Jinqiang Yu \hspace{0.6cm} Alexey Ignatiev \hspace{0.6cm} Peter J. Stuckey \\
  Department of Data Science and AI, Faculty of IT\\
  Monash University, Melbourne, Victoria, Australia \\
  \{\mailtodomain{jinqiang.yu}\texttt{,}\mailtodomain{alexey.ignatiev}\texttt{,}\mailtodomain{peter.stuckey}\}\texttt{@monash.edu} \\
}

\newtheorem{proposition}{Proposition}

\newtheorem{example}{Example}

\newtheorem*{summary*}{Summary}

\declaretheoremstyle[
headpunct={},
headfont=\bfseries,
notefont=\bfseries,
bodyfont=\normalfont,
headformat=\NAME~\NUMBER:\NOTE.
]{def}
\declaretheorem[style=def]{definition}

\newcommand{\fml}[1]{{\mathcal{#1}}}

\newcommand{\mbf}[1]{\ensuremath\mathbf{#1}}
\newcommand{\mbb}[1]{\ensuremath\mathbb{#1}}

\newcommand{\ignore}[1]{}

\newcommand{\ffa}{\textrm{ffa}}
\newcommand{\ffalabel}[1]{\textrm{FFA}$_{\textrm{#1}}$}
\newcommand{\wffalabel}[1]{\textrm{WFFA}$_{\textrm{#1}}$}

\newcommand{\wffa}{\textrm{wffa}}
\newcommand{\axps}{\ensuremath\mbb{A}}
\newcommand{\cxps}{\ensuremath\mbb{C}}

\DeclareMathOperator*{\limply}{\rightarrow}

\def\squareforqed{\hbox{\rlap{$\sqcap$}$\sqcup$}}
\def\qed{\ifmmode\squareforqed\else{\unskip\nobreak\hfil
\penalty50\hskip1em\null\nobreak\hfil\squareforqed
\parfillskip=0pt\finalhyphendemerits=0\endgraf}\fi}

\begin{document}

\maketitle

\begin{abstract}\label{sec:abs}
  Recent years have witnessed the widespread use of artificial
  intelligence (AI) algorithms and machine learning (ML) models.
  Despite their tremendous success, a number of vital problems like ML
  model brittleness, their fairness, and the lack of interpretability
  warrant the need for the active developments in explainable
  artificial intelligence (XAI) and formal ML model verification.
  The two major lines of work in XAI include \emph{feature selection}
  methods, e.g.\ Anchors, and \emph{feature attribution} techniques,
  e.g.\ LIME and SHAP.
  Despite their promise, most of the existing feature selection and
  attribution approaches are susceptible to a range of critical
  issues, including explanation unsoundness and
  \emph{out-of-distribution} sampling.
  A recent formal approach to XAI (FXAI) although serving as an
  alternative to the above and free of these issues suffers from a few
  other limitations.
  For instance and besides the scalability limitation, the formal
  approach is unable to tackle the feature attribution problem.
  Additionally, a formal explanation despite being formally sound is
  typically quite large, which hampers its applicability in practical
  settings.
  Motivated by the above, this paper proposes a way to apply the
  apparatus of formal XAI to the case of feature attribution based on
  formal explanation enumeration.
  Formal feature attribution (FFA) is argued to be advantageous over
  the existing methods, both formal and non-formal.
  Given the practical complexity of the problem, the paper then
  proposes an efficient technique for approximating exact FFA.
  %
  %
  Finally, it offers experimental evidence of the effectiveness of the
  proposed approximate FFA
  %
  %
  in comparison to the existing feature attribution algorithms not
  only in terms of feature importance and but also in terms of their
  relative order.\footnote{Source code and complete experimental setup
  are available at \url{https://github.com/ffattr/ffa.git}.}
  %
  %
\end{abstract}

\section{Introduction} \label{sec:intro}


Thanks to the unprecedented fast growth and the tremendous success,
Artificial Intelligence (AI) and Machine Learning (ML) have become a
universally acclaimed standard in automated decision making causing a
major disruption in computing and the use of technology in
general~\cite{bengio-nature15,jordan-science15,silver-nature15,taward18}.
An ever growing range of practical applications of AI and ML, on the
one hand, and a number of critical issues observed in modern AI
systems (e.g. decision bias~\cite{propublica16} and
brittleness~\cite{szegedy-iclr14}), on the other hand, gave rise to
the quickly advancing area of theory and practice of Explainable AI
(XAI).

Numerous methods exist to explain decisions made by what is called
black-box ML models~\cite{miller-aij19,molnar-bk20}.
Here, \emph{model-agnostic} approaches based on random sampling
prevail~\cite{miller-aij19}, with the most popular being \emph{feature
selection}~\cite{guestrin-aaai18} and \emph{feature
attribution}~\cite{lundberg-nips17,guestrin-aaai18} approaches.
Despite their promise, model-agnostic approaches are susceptible to a
range of critical issues, like unsoundness of
explanations~\cite{ignatiev-ijcai20,huang-corr23} and
\emph{out-of-distribution
sampling}~\cite{lakkaraju-aies20a,lakkaraju-aies20b}, which exacerbates
the problem of trust in AI.

An alternative to model-agnostic explainers is represented by the
methods building on the success of formal reasoning applied to the
logical representations of ML
models~\cite{darwiche-ijcai18,msi-aaai22}.
Aiming to address the limitations of model-agnostic approaches, formal
XAI (FXAI) methods themselves suffer from a few downsides, including
the lack of scalability and the requirement to build a complete logical
representation of the ML model. Formal explanations also tend to be
larger than their model-agnostic counterparts because they do not
reason about (unknown) data distribution~\cite{kutyniok-jair21}.
Finally and most importantly, FXAI methods have not been applied so
far to answer feature attribution questions.

Motivated by the above, we define a novel formal approach to feature
attribution, which builds on the success of existing FXAI
methods~\cite{msi-aaai22}.
By exhaustively enumerating all formal explanations, we can give a
crisp definition of \emph{formal feature attribution} (FFA) as the
proportion of explanations in which a given feature occurs.
We argue that formal feature attribution is hard for the second level
of the polynomial hierarchy.
Although it can be challenging to compute exact FFA in practice, we
show that existing anytime formal explanation enumeration methods can
be applied to efficiently approximate FFA.
Our experimental results demonstrate the effectiveness of the proposed
approach in practice and its advantage over SHAP and LIME given
publicly available tabular and image datasets, as well as on a real
application of XAI in the domain of Software
Engineering~\cite{mcintosh2017fix,pornprasit2021pyexplainer}.

\section{Background} \label{sec:prelim}
%
%
%
This section briefly overviews the status quo in XAI and background
knowledge the paper builds on.

\subsection{Classification Problems}

Classification problems consider a set of classes $\fml{K} = \{1, 2,
\ldots, k\}$\footnote{Any set of classes $\{c_1,\ldots,c_k\}$ can
always be mapped into the set of the corresponding indices
$\{1,\ldots,k\}$.}, and a set of features $\fml{F}=\{1, \ldots, m\}$.
The value of each feature $i \in \fml{F}$ is taken from a domain
$\mbb{D}_i$, which can be categorical or ordinal, i.e.\ integer,
real-valued or Boolean.
%
%
Therefore, the complete feature space is defined as
$\mathbb{F}\triangleq\prod_{i=1}^{m}\mbb{D}_i$.
A concrete point in feature space is represented by
$\mbf{v}=(v_1,\ldots,v_m)\in\mbb{F}$, where each component $v_i \in
\mbb{D}_i$ is a constant taken by feature $i\in\fml{F}$.
An \emph{instance} or \emph{example} is denoted by a specific point
$\mbf{v} \in\mbb{F}$ in feature space and its corresponding class $c
\in \fml{K}$, i.e. a pair $(\mbf{v}, c)$ represents an instance.
Additionally, the notation $\mbf{x} = (x_1,\ldots,x_m)$ denotes an
arbitrary point in feature space, where each component $x_i$ is a
variable taking values from its corresponding domain $\mbb{D}_i$ and
representing feature $i\in\fml{F}$.
A classifier defines a non-constant classification function $\kappa:
\mathbb{F} \limply \fml{K}$.
%

Many ways exist to learn classifiers $\kappa$ given training data,
i.e. a collection of labeled instances $(\mbf{v}, c)$, including
decision trees~\cite{rivest-ipl76} and their
ensembles~\cite{Breiman01,guestrin-kdd16a}, decision
lists~\cite{rivest-ml87}, neural networks~\cite{bengio-nature15}, etc.
Hereinafter, this paper considers boosted tree (BT) models trained
with the use of XGBoost~\cite{guestrin-kdd16a}.

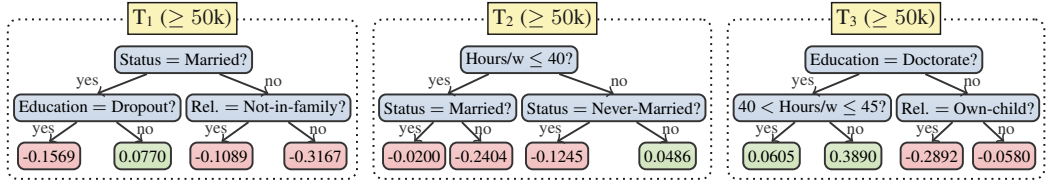
\begin{figure*}[t!]
  \begin{center}
    \scalebox{0.92}{
      \begin{minipage}{0.99\textwidth}
%
%
%

\tikzstyle{box} = [draw=black!90, thick, rectangle, rounded corners,
                     inner sep=10pt, inner ysep=20pt, dotted
                  ]
\tikzstyle{title} = [draw=black!90, fill=black!5, semithick, top color=white,
                     bottom color = black!5, text=black!90, rectangle,
                     font=\small, inner sep=2pt, minimum height=1.3em,
                     top color=tyellow2!27, bottom color=tyellow2!27
                    ]
\tikzstyle{feature} = [rectangle,font=\scriptsize,rounded corners=1mm,thick,%
                       draw=black!80, top color=tblue2!20,bottom color=tblue2!25,%
                       draw, minimum height=1.1em, text centered,%
                       inner sep=2pt%
                      ]
\tikzstyle{pscore} = [rectangle,font=\scriptsize,rounded corners=1mm,thick,%
                     draw=black!80, top color=tgreen3!20,bottom color=tgreen3!27,%
                     draw, minimum height=1.1em, text centered,%
                     inner sep=2pt%
                    ]
\tikzstyle{nscore} = [rectangle,font=\scriptsize,rounded corners=1mm,thick,%
                     draw=black!80, top color=tred2!20,bottom color=tred2!25,%
                     draw, minimum height=1.1em, text centered,%
                     inner sep=2pt%
                    ]

\begin{adjustbox}{center}
\setlength{\tabcolsep}{3pt}
\def\arraystretch{3}
\begin{tabular}{ccc}
    \begin{tikzpicture}[node distance = 4.0em, auto]
        \node [box] (box) {%
        \begin{minipage}[t!]{0.314\textwidth}
            \vspace{0.9cm}\hspace{1.5cm}
        \end{minipage}
        };
        \node[title] at (box.north) {$\text{T}_\text{1}$ ($\geq 50$k)};

        \node [feature] (feat1) at (0.016, 0.57) {Status $=$ Married?};
        \node [feature, below left  = 0.8em and -2.78em of feat1] (feat2) {Education $=$ Dropout?};
        \node [feature, below right = 0.8em and -2.78em of feat1] (feat3) {Rel. $=$ Not-in-family?};

        \node [nscore, below left  = 0.8em and -2.8em of feat2] (pos1) {-0.1569};
        \node [pscore, below right  = 0.8em and -2.8em of feat2] (neg1) {0.0770};
        \node [nscore, below left = 0.8em and -2.8em of feat3] (pos2) {-0.1089};
        \node [nscore, below right = 0.8em and -2.8em of feat3] (neg2) {-0.3167};

        \draw [->,thick,black!80] (feat1) to[] node[above, pos=1.2, font=\scriptsize] {yes} (feat2.north);
        \draw [->,thick,black!80] (feat1) to[] node[above, pos=1.1, font=\scriptsize] { no} (feat3.north);
        \draw [->,thick,black!80] (feat2) to[] node[above, pos=1.2, font=\scriptsize] {yes} (pos1.north);
        \draw [->,thick,black!80] (feat2) to[] node[above, pos=1.1, font=\scriptsize] { no} (neg1.north);
        \draw [->,thick,black!80] (feat3) to[] node[above, pos=1.2, font=\scriptsize] {yes} (pos2.north);
        \draw [->,thick,black!80] (feat3) to[] node[above, pos=1.1, font=\scriptsize] { no} (neg2.north);
    \end{tikzpicture}
    &
    \begin{tikzpicture}[node distance = 4.0em, auto]
        \node [box] (box) {%
        \begin{minipage}[t!]{0.302\textwidth}
            \vspace{0.9cm}\hspace{1.5cm}
        \end{minipage}
        };
        \node[title] at (box.north) {$\text{T}_\text{2}$ ($\geq 50$k)};

        \node [feature] (feat1) at (-0.335, 0.57) {Hours/w $\leq$ 40?};
        \node [feature, below left  = 0.8em and -2.36em of feat1] (feat2) {Status $=$ Married?};
        \node [feature, below right = 0.8em and -2.36em of feat1] (feat3) {Status $=$ Never-Married?};

        \node [nscore, below left  = 0.8em and -2.8em of feat2] (pos1) {-0.0200};
        \node [nscore, below right  = 0.8em and -2.8em of feat2] (neg1) {-0.2404};
        \node [nscore, below left = 0.8em and -2.8em of feat3] (pos2) {-0.1245};
        \node [pscore, below right = 0.8em and -2.8em of feat3] (neg2) {0.0486};

        \draw [->,thick,black!80] (feat1) to[] node[above, pos=1.2, font=\scriptsize] {yes} (feat2.north);
        \draw [->,thick,black!80] (feat1) to[] node[above, pos=1.1, font=\scriptsize] { no} (feat3.north);
        \draw [->,thick,black!80] (feat2) to[] node[above, pos=1.2, font=\scriptsize] {yes} (pos1.north);
        \draw [->,thick,black!80] (feat2) to[] node[above, pos=1.1, font=\scriptsize] { no} (neg1.north);
        \draw [->,thick,black!80] (feat3) to[] node[above, pos=1.2, font=\scriptsize] {yes} (pos2.north);
        \draw [->,thick,black!80] (feat3) to[] node[above, pos=1.1, font=\scriptsize] { no} (neg2.north);
    \end{tikzpicture}
    &
     \begin{tikzpicture}[node distance = 4.0em, auto]
        \node [box] (box) {%
        \begin{minipage}[t!]{0.28\textwidth}
            \vspace{0.9cm}\hspace{1.5cm}
        \end{minipage}
        };
        \node[title] at (box.north) {$\text{T}_\text{3}$ ($\geq 50$k)};

        \node [feature] (feat1) at (0.113, 0.57) {Education $=$ Doctorate?};
        \node [feature, below left  = 0.8em and -3.6em of feat1] (feat2) {40 $<$ Hours/w $\leq$ 45?};
        \node [feature, below right = 0.8em and -3.6em of feat1] (feat3) {Rel. $=$ Own-child?};

        \node [pscore, below left  = 0.8em and -2.8em of feat2] (pos1) {0.0605};
        \node [pscore, below right  = 0.8em and -2.8em of feat2] (neg1) {0.3890};
        \node [nscore, below left = 0.8em and -2.8em of feat3] (pos2) {-0.2892};
        \node [nscore, below right = 0.8em and -2.8em of feat3] (neg2){-0.0580};

        \draw [->,thick,black!80] (feat1) to[] node[above, pos=1.2, font=\scriptsize] {yes} (feat2.north);
        \draw [->,thick,black!80] (feat1) to[] node[above, pos=1.1, font=\scriptsize] { no} (feat3.north);
        \draw [->,thick,black!80] (feat2) to[] node[above, pos=1.2, font=\scriptsize] {yes} (pos1.north);
        \draw [->,thick,black!80] (feat2) to[] node[above, pos=1.1, font=\scriptsize] { no} (neg1.north);
        \draw [->,thick,black!80] (feat3) to[] node[above, pos=1.2, font=\scriptsize] {yes} (pos2.north);
        \draw [->,thick,black!80] (feat3) to[] node[above, pos=1.1, font=\scriptsize] { no} (neg2.north);
    \end{tikzpicture}
    \\
   \end{tabular}
\end{adjustbox}

      \end{minipage}
    }
  \end{center}
  \caption{Example boosted tree model~\cite{guestrin-kdd16a} trained
  on the well-known \emph{adult} classification dataset.}
  \label{fig:bt}
\end{figure*}

\begin{figure}[t]
  \centering
      \begin{subfigure}[b]{0.225\textwidth}
    \centering
    \raisebox{0.2cm}{
      \includegraphics[width=\textwidth]{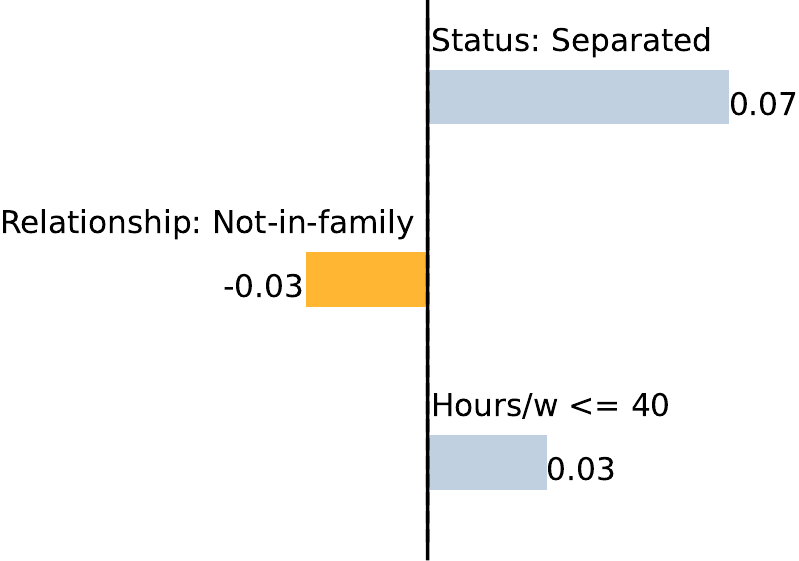}
    }
    \caption{LIME}
    \label{fig:lime-example}
   \end{subfigure}%
   \hfill
   \begin{subfigure}[b]{0.23\textwidth}
    \centering
    \raisebox{0.2cm}{
      \includegraphics[width=\textwidth]{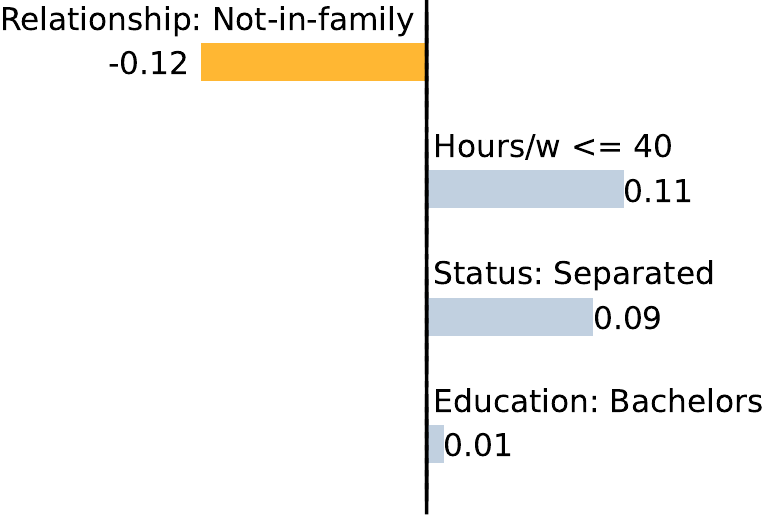}
    }
    \caption{SHAP}
    \label{fig:shap-example}
   \end{subfigure}%
   \hfill
   \begin{subfigure}[b]{0.25\textwidth}
    \centering
    \tiny
    \setlength{\tabcolsep}{2pt}
    \def\arraystretch{0.9}
        \begin{tabular}{ll}  \toprule
                \multicolumn{2}{c}{\bf $\bm{\fml{X}_{1}=}$ \{\,Education, Hours/w\,\}} \\ \midrule
		IF & Education $=$ Bachelors \\
		AND & Hours/w $\leq$ 40 \\
		THEN & Target $<$50$k$ \\ \midrule
                \multicolumn{2}{c}{\bf $\bm{\fml{X}_{2}=}$ \{\,Education, Status\,\}} \\ \midrule
		IF & Education $=$ Bachelors \\
		AND & Status $=$ Separated \\
		THEN & Target $<$50$k$ \\ \bottomrule
        \end{tabular}
    \caption{AXp's $\fml{X}_1$ and $\fml{X}_2$}
    \label{fig:exaxp}
   \end{subfigure}%
   \hfill
   \begin{subfigure}[b]{0.145\textwidth}
    \centering
    \raisebox{0.2cm}{
      \includegraphics[width=\textwidth]{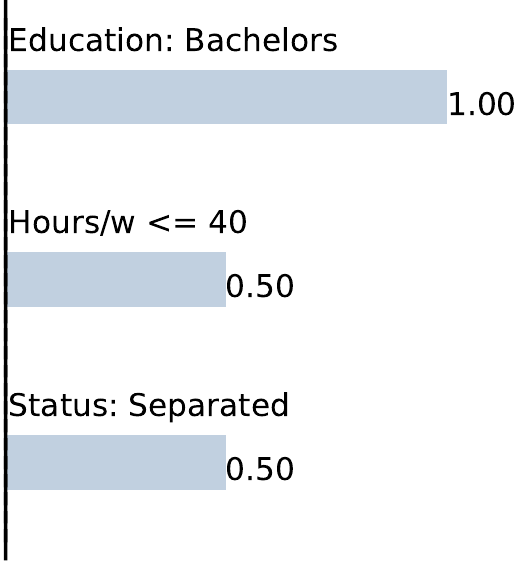}
    }
    \caption{FFA}
    \label{fig:ffa-example}
   \end{subfigure}%
   \caption{Examples of feature attribution reported by LIME and SHAP,
   as well as both AXp's (no more AXp's exist) followed by FFA for the instance $\mbf{v}$ shown
   in \autoref{ex:classifier}.}
    \label{fig:explex}
\end{figure}

\begin{example} \label{ex:classifier}
  \autoref{fig:bt} shows a BT model trained for a
  simplified version of the \emph{adult} dataset~\cite{kohavi-kdd96}.
  For a data instance $\mbf{v} =
  \{\text{Education}=\text{Bachelors},$
  $\text{Status}=\text{Separated}$, $\text{Occupation}=\text{Sales}$,
  Relation\-ship~$=\text{Not-in-family}, \text{Sex}=\text{Male},$
  $\text{Hours/w}\leq\text{40}\}$, the model predicts $<$50$k$
  because the sum of the weights in the 3 trees for this
  instance equals $-0.4073=(-0.1089 -0.2404 -0.0580)<0$.
\end{example}

\subsection{ML Model Interpretability and Post-Hoc Explanations}
Interpretability is generally accepted to be a subjective concept,
without a formal definition~\cite{lipton-cacm18}.
One way to measure interpretability is in terms of the succinctness of
information provided by an ML model to justify a given prediction.
Recent years have witnessed an upsurge in the interest in devising and
applying interpretable models in safety-critical
applications~\cite{rudin-natmi19,molnar-bk20}.
An alternative to interpretable models is post-hoc explanation of
\emph{black-box} models, which this paper focuses on.

Numerous methods to compute explanations have been proposed
recently~\cite{miller-aij19,molnar-bk20}.
The lion's share of these comprise what is called
\emph{model-agnostic} approaches to
explainability~\cite{guestrin-kdd16,lundberg-nips17,guestrin-aaai18}
of heuristic nature that resort to extensive sampling in the vicinity
of an instance being explained in order to ``estimate'' the behavior
of the classifier in this local vicinity of the instance.
In this regard, they rely on estimating input data distribution by
building on the information about the training
data~\cite{lakkaraju-aies20b}.
Depending on the form of explanations model-agnostic approaches offer,
they are conventionally classified as \emph{feature selection} or
\emph{feature attribution} approaches briefly discussed below.

\paragraph{Feature Selection.}
A feature selection approach identifies subsets of features that are
deemed \emph{sufficient} for a given prediction $c=\kappa(\mbf{v})$.
As mentioned above, the majority of feature selection approaches are
model-agnostic with one prominent example being
Anchors~\cite{guestrin-aaai18}.
As such, the sufficiency of the selected set of features for a given
prediction is determined statistically based on extensive sampling
around the instance of interest, by assessing a few measures like
\emph{fidelity}, \emph{precision}, among others.
As a result, feature selection explanations given as a set of features
$\omega\subseteq\fml{F}$ should be interpreted as the conjunction
$\bigwedge_{i\in\omega}{(x_i=v_i)}$ deemed responsible for prediction
$c=\kappa(\mbf{v})$, $\mbf{v}\in\mbb{F}$, $c\in\fml{K}$.
Due to the statistical nature of these explainers, they are known to
suffer from various explanation quality
issues~\cite{ignatiev-ijcai20,lakkaraju-aies20b,lakkaraju-nips21}.
An additional line of work on \emph{formal}
explainability~\cite{darwiche-ijcai18,inms-aaai19} also tackles
feature selection while offering guarantees of soundness; these are
discussed below.

\paragraph{Feature Attribution.}
A different view on post-hoc explanations is provided by feature
attribution approaches, e.g.\ LIME~\cite{guestrin-kdd16} and
SHAP~\cite{lundberg-nips17}.
Based on random sampling in the neighborhood of the target instance,
these approaches attribute responsibility to all model's features by
assigning a numeric value $w_i\in\mbb{R}$ of importance to each
feature $i\in\fml{F}$.
Given these importance values, the features can then be ranked from
most important to least important.
As a result, a feature attribution explanation is conventionally
provided as a linear form $\sum_{i\in\fml{F}}{w_i\cdot x_i}$, which
can be also seen as approximating the original black-box explainer
$\kappa$ in the \emph{local} neighborhood of instance
$\mbf{v}\in\mbb{F}$.
Among other feature attribution approaches,
SHAP~\cite{lundberg-nips17,barcelo-aaai21,barcelo-corr21} is often
claimed to stand out as it aims at approximating Shapley values, a
powerful concept originating from cooperative games in game
theory~\cite{shapley-ctg53}.

\paragraph{Formal Explainability.}
In this work, we build on formal explainability proposed in earlier
work~\cite{darwiche-ijcai18,inms-aaai19,darwiche-ecai20,marquis-kr20,msi-aaai22}.
where explanations are equated with \emph{abductive explanations}
(AXp's).
Abductive explanations are \emph{subset-minimal} sets of features
formally proved to suffice to explain an ML prediction given a
formal representation of the classifier of interest.
Concretely, given an instance $\mbf{v} \in \mbb{F}$ and a prediction
$c=\kappa(\mbf{v})$, an AXp is a subset-minimal set of features
$\fml{X} \subseteq \fml{F}$, such that
\begin{equation} \label{eq:axp}
  \forall(\mbf{x} \in \mbb{F}). \bigwedge\nolimits_{i \in \fml{X}}
  (x_i = v_i) \limply (\kappa(\mbf{x}) = c)
\end{equation}
%


Abductive explanations are guaranteed to be subset-minimal sets of
features proved to satisfy~\eqref{eq:axp}.
As other feature selection explanations, they answer \emph{why} a
certain prediction was made.
An alternate way to explain a model's behavior is to seek an answer
\emph{why not} another prediction was made, or, in other words,
\emph{how} to change the prediction.
Explanations answering \emph{why not} questions are referred to as
\emph{contrastive explanations}
(CXp's)~\cite{miller-aij19,inams-aiia20,msi-aaai22}.
As in prior work, we define a CXp as a subset-minimal set of features
that, if allowed to change their values, are \emph{necessary} to
change the prediction of the model.
Formally, a CXp for prediction $c = \kappa(\mbf{v})$ is a
subset-minimal set of features $\fml{Y}\subseteq\fml{F}$, such that
\begin{equation} \label{eq:cxp}
  \exists(\mbf{x}\in\mbb{F}).\bigwedge\nolimits_{i\not\in\fml{Y}}(x_i=v_i)\land(\kappa(\mbf{x})\not=c)
\end{equation}


Finally, recent work has shown that AXp's and CXp's for a given
instance $\mbf{v}\in\mbb{F}$ are related through the \emph{minimal
hitting set duality}~\cite{inams-aiia20,reiter-aij87}.
The duality implies that each AXp for a prediction $c=\kappa(\mbf{v})$
is a \emph{minimal hitting set}\footnote{Given a set of sets
  $\mbb{S}$, a \emph{hitting set} of $\mbb{S}$ is a set $H$ such that
  $\forall S \in \mbb{S}, S \cup H \neq \emptyset$, i.e. $H$ ``hits''
every set in $\mbb{S}$. A hitting set $H$ for $\mbb{S}$ is
\emph{minimal} if none of its strict subsets is also a hitting set.}
(MHS) of the set of all CXp's for that prediction, and the other way
around: each CXp is an MHS of the set of all AXp's.
The explanation enumeration algorithm~\cite{inams-aiia20} applied in
this paper heavily relies on this duality relation and is inspired by
the MARCO algorithm originating from the area of over-constrained
systems~\cite{lm-cpaior13,pms-aaai13,lpmms-cj16}.
A growing body of recent work on formal explanations is represented
(but not limited)
by~\cite{msgcin-nips20,msgcin-icml21,ims-ijcai21,ims-sat21,barcelo-nips21,kutyniok-jair21,darwiche-jair21,kwiatkowska-ijcai21,mazure-cikm21,tan-nips21,iims-jair22,rubin-aaai22,iisms-aaai22,hiicams-aaai22,msi-aaai22,an-ijcai22,leite-kr22,jpms-rw22,barcelo-nips22,jpms-corr22,yisnms-aaai23,jpms-aaai23,jpms-aij23,msi-fai23,ihincms-ajar23,izza-corr23,jpms-corr23,huang-corr23b}.

\begin{example} \label{ex:expls}
  In the context of \autoref{ex:classifier}, feature attribution
  computed by LIME and SHAP as well as all 2 AXp's are shown in
  \autoref{fig:explex}.
  %
  %
  %
  AXp $\fml{X}_1$ indicates that specifying Education $=$ Bachelors and
  Hours/w $\leq$ 40 guarantees that any compatible instance is
  classified as $<$ 50k independent of the values of other features,
  e.g. Status and Relationship, since the maximal sum of weights is
  $0.0770 -0.0200 -0.0580 = -0.0010 < 0$ as long as the feature values
  above are used.
  %
  %
  Observe that another AXp $\fml{X}_2$ for $\mbf{v}$ is \{Education, Status\}.
  %
  Since both of the two AXp's for $\mbf{v}$ consist of two features,
  it is difficult to judge which one is better without a formal
  feature importance assessment.
\end{example}

\section{Why Formal Feature Attribution?} \label{sec:fattr}

On the one hand, abductive explanations serve as a viable alternative
to non-formal feature selection approaches because they (i)~guarantee
subset-minimality of the selected sets of features and (ii)~are
computed via formal reasoning over the behavior of the corresponding
ML model.
Having said that, they suffer from a few issues.
First, observe that deciding the validity of~\eqref{eq:axp} requires a
formal reasoner to take into account the complete feature space
$\mbb{F}$, assuming that the features are independent and uniformly
distributed~\cite{kutyniok-jair21}.
In other words, the reasoner has to check all the combinations of
feature values, including those that \emph{never appear in practice}.
This makes AXp's being unnecessarily \emph{conservative} (long), i.e.\
they may be hard for a human decision maker to interpret.
Second, AXp's are not aimed at providing feature attribution.
The abundance of various AXp's for a single data
instance~\cite{inms-aaai19}, e.g. see~\autoref{ex:expls}, exacerbates
this issue as it becomes unclear for a user which of the AXp's to use
to make an informed decision in a particular situation.

On the other hand, non-formal feature attribution in general is known
to be susceptible to out-of-distribution
sampling~\cite{lakkaraju-aies20b,lakkaraju-aies20a} while SHAP is
shown to fail to effectively approximate Shapley
values~\cite{huang-corr23}.
Moreover and quite surprisingly,~\cite{huang-corr23} argued that even
the use of exact Shapley values is inadequate as a measure of feature
importance.
Our results below confirm that both LIME and SHAP often fail to grasp
the real feature attribution in a number of practical scenarios.


To address the above limitations, we propose the concept of
\emph{formal feature attribution} (FFA) as defined next.
(An insight on this was also given in~\cite{huang-corr23}.)
Let us denote the set of all formal abductive explanations for a
prediction $c=\kappa(\mbf{v})$ by $\axps_\kappa(\mbf{v}, c)$.
%
%
Then formal feature attribution of a feature $i\in\fml{F}$ can be
defined as the proportion of abductive explanations where it occurs.
More formally,

\begin{definition}[FFA]
  The \emph{formal feature attribution} $\ffa_\kappa(i,(\mbf{v},c))$
  of a feature $i \in \fml{F}$ to an instance $(\mbf{v}, c)$ for
  machine learning model $\kappa$ is
  \begin{equation} \label{eq:ffa}
    \ffa_\kappa(i,(\mbf{v}, c)) = \frac{|\{ \fml{X} ~|~ \fml{X} \in
      \axps_\kappa(\mbf{v}, c), i \in \fml{X}) |}{|
    \axps_\kappa(\mbf{v}, c)|}
  \end{equation}
\end{definition}

Formal feature attribution has some nice properties.
First, it has a strict and formal definition, i.e.\ we can, assuming
we are able to compute the complete set of AXp's for an instance,
exactly define it for all features $i \in \fml{F}$.
Second, it is fairly easy to explain to a user of the classification
system, even if they are non-expert.
Namely, it is the percentage of (formal abductive) explanations that
make use of a particular feature $i$.
Third, as we shall see later, even though we may not be able to
compute all AXp's exhaustively, we can still get good approximations
fast.

\begin{example} \label{ex:ffa}
  Recall \autoref{ex:expls}.
  As there are 2 AXp's for instance $\mbf{v}$, the prediction can be
  attributed to the 3 features with non-zero FFA shown in
  \autoref{fig:ffa-example}.
  Also, observe how both LIME and SHAP (see \autoref{fig:lime-example}
  and \autoref{fig:shap-example}) assign non-zero attribution to the
  feature Relationship, which is in fact irrelevant for the
  prediction, but overlook the highest importance of feature
  Education.
\end{example}

One criticism of the above definition is that it does not take into
account the length of explanations where the feature arises.
Arguably if a feature arises in many AXp's of size 2, it should be
considered more important than a feature which arises in the same
number of AXp's but where each is of size 10.
An alternate definition, which tries to take this into account, is the
weighted formal feature attribution (WFFA), i.e.\ the \emph{average}
proportion of AXp's that include feature $i\in\fml{F}$.
Formally,
\begin{definition}[WFFA]
  The \emph{weighted formal feature attribution}
  $\wffa_\kappa(i,(\mbf{v},c))$ of a feature $i \in \fml{F}$ to an
  instance $(\mbf{v}, c)$ for machine learning model $\kappa$ is
  \begin{equation} \label{eq:wffa}
    \wffa_\kappa(i,(\mbf{v}, c)) = \frac{ \sum_{\fml{X} \in
    \axps_\kappa(\mbf{v}, c), i \in \fml{X}}
  |\fml{X}|^{-1}}{| \axps_\kappa(\mbf{v}, c)|}
\end{equation}
\end{definition}

Note that these attribution values are not on the same scale although
they are convertible:

$$
\sum_{i \in
  \fml{F}} \ffa_\kappa(i,(\mbf{v},
  c)) =  \frac{\sum_{\fml{X} \in
  \axps_\kappa(\mbf{v}, c)}|\fml{X}|}{| \axps_\kappa(\mbf{v}, c)|} \times
  \sum_{i \in
  \fml{F}} \wffa_\kappa(i,(\mbf{v},
  c)).
$$

FFA can be related to the problem of \emph{feature
relevancy}~\cite{huang-tacas23}, where a feature is said to be
\emph{relevant} if it belongs to at least one AXp.
Indeed, feature $i\in\fml{F}$ is relevant for prediction
$c=\kappa(\mbf{v})$ if and only if $\ffa_\kappa(i, (\mbf{v},c))>0$.
As a result, the following claim can be made.

\begin{proposition} \label{prop:complexity}
  Given a feature $i\in\fml{F}$ and a prediction $c=\kappa(\mbf{v})$,
  deciding whether $\ffa_\kappa(i, (\mbf{v}, c))>\omega$,
  $\omega\in(0,1]$, is at least as hard as deciding whether feature
  $i$ is relevant for the prediction.
\end{proposition}

The above result indicates that computing exact FFA values may be
expensive in practice.
For example and in light of~\cite{huang-tacas23}, one can conclude
that the decision version of the problem is $\Sigma_2^\text{P}$-hard
in the case of DNF classifiers.

Similarly and using the relation between FFA and feature relevancy
above, we can note that the decision version of the problem is in
$\Sigma_2^\text{P}$ as long as deciding the validity of~\eqref{eq:axp}
is in NP, which in general is the case (unless the problem is simpler,
e.g. for decision trees~\cite{iims-jair22}).
Namely, the following result is a simple consequence of the membership
result for the feature relevance problem~\cite{huang-tacas23}.

\begin{proposition} \label{prop:membership}
  Deciding whether $\ffa_\kappa(i, (\mbf{v}, c))>0$ is in
  $\Sigma_2^\text{P}$ if deciding~\eqref{eq:axp} is in NP.
\end{proposition}

\section{Approximating Formal Feature Attribution} \label{sec:approx}

As the previous section argues and as our experimental results
confirm, it may be challenging in practice to compute exact FFA values
due to the general complexity of the problem.
Although some ML models admit efficient formal encodings and reasoning
procedures, effective principal methods for FFA approximation seem
necessary.
This section proposes one such method.

Normally, formal explanation enumeration is done by exploiting the MHS
duality between AXp's and CXp's and the use of
MARCO-like~\cite{lpmms-cj16} algorithms aiming at efficient
exploration of minimal hitting sets of either AXp's or
CXp's~\cite{lm-cpaior13,pms-aaai13,lpmms-cj16,inams-aiia20}.
Depending on the target type of formal explanation, MARCO exhaustively
enumerates all such explanations one by one, each time extracting a
candidate minimal hitting set and checking if it is a desired
explanation.
If it is then it is recorded and blocked such that this candidate is
never repeated again.
Otherwise, a dual explanation is extracted from the subset of features
complementary to the candidate~\cite{inms-aaai19}, gets recorded and
blocked so that it is hit by each future candidate.
The procedure proceeds until no more hitting sets of the set of dual
explanations can be extracted, which signifies that all target
explanations are enumerated.
Observe that while doing so, MARCO also enumerates all the dual
explanations as a kind of ``side effect''.

One of the properties of MARCO used in our approximation approach is
that it is an \emph{anytime} algorithm, i.e.\ we can run it for as
long as we need to get a sufficient number of explanations.
This means we can stop it by using a timeout or upon collecting a
certain number of explanations.

The main insight of FFA approximation is as follows.
Recall that to compute FFA, we are interested in AXp enumeration.
Although intuitively this suggests the use of MARCO targeting AXp's,
for the sake of fast and high-quality FFA approximation, we propose to
target CXp enumeration with AXp's as dual explanations computed
``unintentionally''.
The reason for this is twofold: (i)~we need to get a good FFA
approximation as fast as we can and (ii)~according to our practical
observations, MARCO needs to amass a large number of dual explanations
before it can start producing target explanations.
This is because the hitting set enumerator is initially ``blind'' and
knows nothing about the features it should pay attention to --- it
uncovers this information gradually by collecting dual explanations to
hit.
This way a large number of dual explanations can quickly be enumerated
during this initial phase of grasping the search space, essentially
``for free''.
Our experimental results demonstrate the effectiveness of this
strategy in terms of monotone convergence of approximate FFA to the
exact FFA with the increase of the time limit.
A high-level view of the version of MARCO used in our approach
targeting CXp enumeration and amassing AXp's as dual explanations is
shown in \autoref{alg:enum}.

\newcommand{\Break}{\textbf{break}}
\algnewcommand{\IfThen}[2]{
  \State \algorithmicif\ #1\ \algorithmicthen\ #2}
\algnewcommand{\IfThenElse}[3]{
  \State \algorithmicif\ #1\ \algorithmicthen\ #2\ \algorithmicelse\ #3}

\begin{algorithm}[t]
  \begin{algorithmic}[1]
    \Procedure{XpEnum}{$\kappa$, $\mbf{v}$, $c$}
      \State $\left(\axps, \cxps\right)\gets (\emptyset, \emptyset)$ \label{ln:init}
      \Comment{Sets of AXp's and CXp's to collect.}
      \While{true}
        \State $\fml{Y}\gets\Call{MinimalHS}{\axps, \cxps}$ \label{ln:mhs}
        \Comment{Get a new MHS of $\axps$ subject to $\cxps$.}
        \IfThen{$\fml{Y}=\bot$}{\Break} \label{ln:nocand}
        \Comment{Stop if none is computed.}
        \If{$\exists(\mbf{x}\in\mbb{F}).\bigwedge\nolimits_{i\not\in\fml{Y}}(x_i=v_i)\land(\kappa(\mbf{x})\not=c)$} \label{ln:check}
          \Comment{Check CXp condition \eqref{eq:cxp} for $\fml{Y}$.}
          \State $\cxps\gets\cxps\cup\{\fml{Y}\}$ \label{ln:cxprec}
          \Comment{$\fml{Y}$ appears to be a CXp.}
        \Else
          \Comment{There must be a missing AXp $\fml{X}\subseteq\fml{F}\setminus\fml{Y}$.}
          \State $\fml{X}\gets\Call{ExtractAXp}{\fml{F}\setminus\fml{Y},\kappa,\mbf{v},c}$ \label{ln:extract}
          \Comment{Get AXp $\fml{X}$ by iteratively checking~\eqref{eq:axp}~\cite{inms-aaai19}.}
          \State $\axps\gets\axps\cup\{\fml{X}\}$ \label{ln:axprec}
          \Comment{Collect new AXp $\fml{X}$.}
        \EndIf
      \EndWhile
      \Return{$\axps$, $\cxps$}
    \EndProcedure
  \end{algorithmic}
  \caption{MARCO-like Anytime Explanation Enumeration}
  \label{alg:enum}
\end{algorithm}


\section{Experimental Evidence} \label{sec:res}

This section assesses the formal feature attribution for gradient
boosted trees~(BT)~\cite{guestrin-kdd16a} on multiple widely used
images and tabular datasets, and compares FFA with LIME and SHAP.
In addition, it also demonstrates the use of FFA in a real-world
scenario of Just-in-Time~(JIT) defect prediction, which assists teams
in prioritizing their limited resources on high-risk commits or pull
requests~\cite{pornprasit2021pyexplainer}.

\paragraph{Setup and Prototype Implementation.}
All experiments were performed on an Intel Xeon 8260 CPU running
Ubuntu 20.04.2 LTS, with the memory limit of 8 GByte.
A prototype of the approach implementing \autoref{alg:enum} and thus
producing FFA was developed as a set of Python scripts and builds
on~\cite{iisms-aaai22}.
%
%
As the FFA and WFFA values turn out to be almost identical (subject to
normalization) in our experiments, here we report only FFA.
WFFA results can be found in supplementary material.

\paragraph{Datasets and Machine Learning Models.}
The well-known MNIST dataset~\cite{deng2012mnist,pytorch-neurips19} of
hand-written digits 0--9 is considered, with two concrete binary
classification tasks created: 1 vs.~3 and 1 vs.~7.
We also consider PneumoniaMNIST~\cite{medmnistv2}, a binary
classification dataset to distinguish X-ray images of pneumonia from
normal cases.
To demonstrate extraction of \emph{exact} FFA values for the above
datasets, we also examine their downscaled versions, i.e.\ reduced
from $\text{28} \times \text{28} \times \text{1}$ to $\text{10} \times
\text{10} \times \text{1}$.
We also consider 11 tabular datasets often applied in the area of ML
explainability and
fairness~\cite{Olson2017PMLB,dua-2019,schmidt-1988,propublica16,fairml17,fairness15}.
All the considered datasets are randomly split into 80\% training and
and 20\% test data.
For images, 15 test instances are randomly selected in each test set
for explanation while all tabular test instances are explained.
%
%
For all datasets, gradient boosted trees~(BTs) are trained by
XGBoost~\cite{guestrin-kdd16a}, where each BT consists of 25 trees of
depth 3 per class.\footnote{Test accuracy for MNIST digits is 0.99,
  while it is 0.83 for PneumoniaMNIST. This holds both for the 28
  $\times$ 28 and 10 $\times$ 10 versions of the datasets.
The average accuracy across the 11 selected tabular datasets is 0.80.
}
%
%
Finally, we show the use of FFA on 2 JIT defect prediction
datasets~\cite{pornprasit2021pyexplainer}, with 500 instances per
dataset chosen for analysis.

\begin{figure}[t]
  \centering
   \begin{subfigure}[b]{0.33\textwidth}
    \centering
    \includegraphics[width=\textwidth]{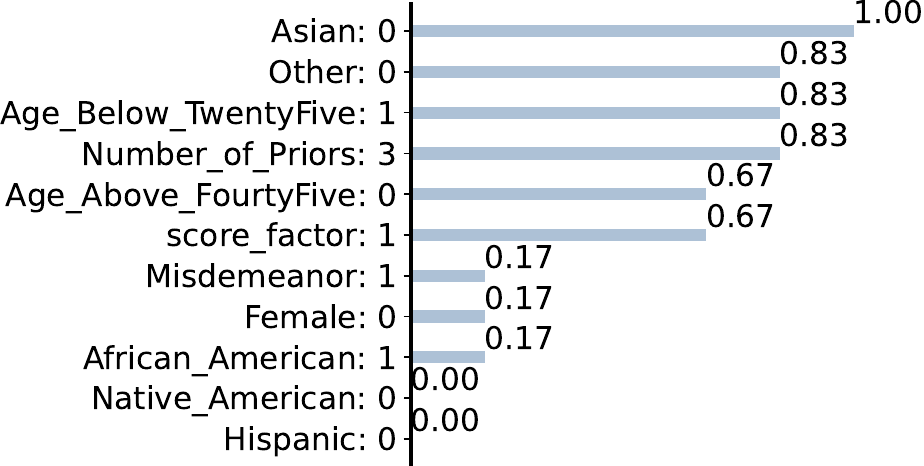}
    \caption{FFA}
    \label{fig:formal-compas}
   \end{subfigure}%
   \begin{subfigure}[b]{0.33\textwidth}
    \centering
    \includegraphics[width=\textwidth]{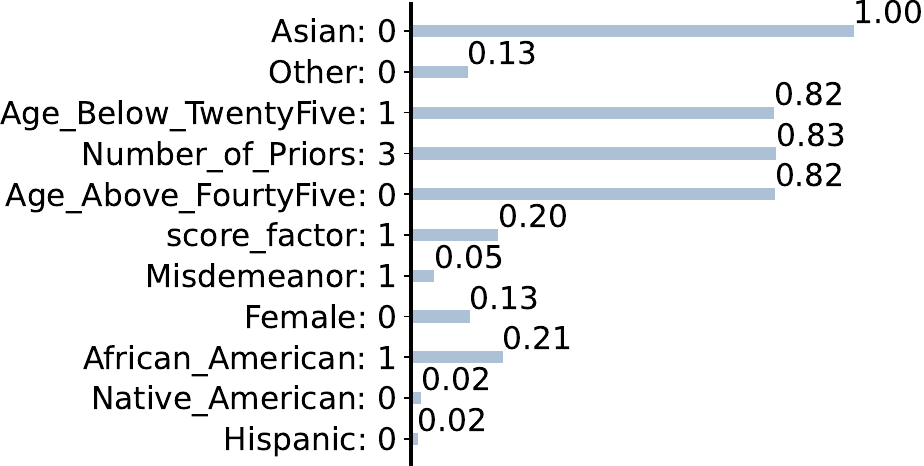}
    \caption{LIME}
    \label{fig:lime-compas}
   \end{subfigure}%
   \begin{subfigure}[b]{0.33\textwidth}
    \centering
    \includegraphics[width=\textwidth]{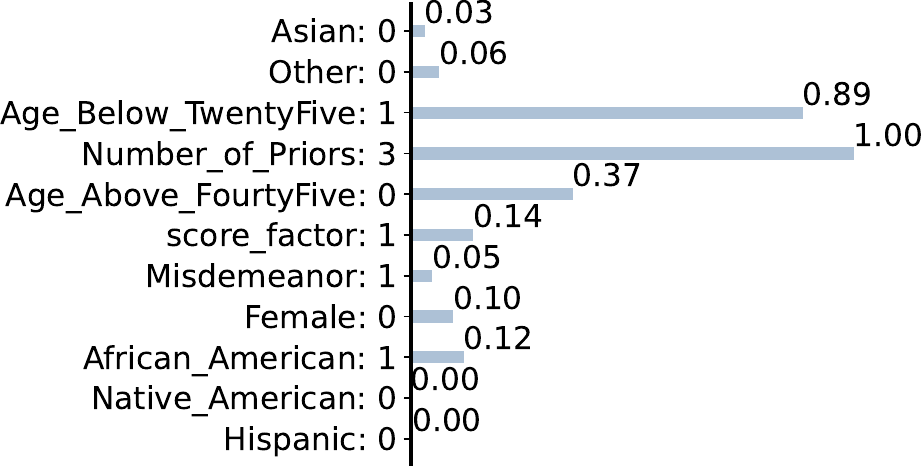}
    \caption{SHAP}
    \label{fig:shap-compas}
   \end{subfigure}%
    \caption{Explanations for an instance of Compas
		$\mbf{v} =  \{
			\text{\#Priors}=\text{3}, \text{Score\_factor}=\text{1},
			\text{Age\_Above\_FourtyFive}=\text{0}, \text{Age\_Below\_TwentyFive}=\text{1},
 			\text{African\_American}=\text{1}, \text{Asian}=\text{0},
			\text{Hispanic}=\text{0}, \text{Native\_American}=\text{0},
			\text{Other}=\text{0}, \text{Female}=\text{0},
			\text{Misdemeanor}=\text{1}
			\}$
	    predicted as Two\_yr\_Recidivism $=$ true.
    }
    \label{fig:compas}
\end{figure}


\subsection{Formal Feature Attribution}\label{sec:ffares}

\begin{table}[t!]
\caption{LIME and SHAP versus FFA on tabular data.
}
\label{tab:tabffa}
\centering
\scalebox{0.77}{
	\setlength{\tabcolsep}{1.6pt}
	\begin{tabular}{cccccccccccc}  \toprule
\textbf{Dataset} & \textbf{adult} & \textbf{appendicitis} & \textbf{australian} & \textbf{cars} & \textbf{compas} & \textbf{heart-statlog} & \textbf{hungarian} & \textbf{lending} & \textbf{liver-disorder} & \textbf{pima} & \textbf{recidivism}  \\
($|\fml{F}|$)  & (12) & (7) & (14) & (8) & (11) & (13) & (13) & (9) & (6) & (8) & (15)  \\ \midrule
 \textbf{Approach} & \multicolumn{11}{c}{\textbf{Error}} \\ \midrule
LIME & 4.48 & 2.25 & 5.13 & 1.53 & 3.28 & 4.48 & 4.56 & 1.39 & 2.39 & 2.72 & 4.73  \\
SHAP & 4.47 & 2.01 & 4.49 & 1.40 & 2.67 & 3.71 & 4.14 & 1.44 & 2.28 & 3.00 & 4.76  \\  \midrule
 & \multicolumn{11}{c}{\textbf{Kendall’s Tau}} \\ \midrule
LIME & 0.07 & 0.11 & 0.22 & -0.11 & -0.11 & 0.17 & 0.04 & -0.36 & -0.22 & 0.17 & 0.05  \\
SHAP & 0.03 & 0.12 & 0.27 & -0.10 & -0.10 & 0.17 & 0.20 & -0.39 & -0.21 & 0.07 & 0.12  \\  \midrule
 & \multicolumn{11}{c}{\textbf{RBO}} \\ \midrule
LIME & 0.54 & 0.66 & 0.49 & 0.63 & 0.55 & 0.56 & 0.41 & 0.59 & 0.66 & 0.68 & 0.39  \\
SHAP & 0.49 & 0.67 & 0.55 & 0.66 & 0.59 & 0.52 & 0.49 & 0.61 & 0.67 & 0.63 & 0.44  \\  \bottomrule
	\end{tabular}
}
\end{table}

In this section, we restrict ourselves to examples where we can
compute the \emph{exact} FFA values for explanations by computing all
AXp's.
To compare with LIME and SHAP, we take their solutions, replace
negative attributions by the positive counterpart (in a sense taking
the absolute value) and then normalize the values into
$[\text{0}, \text{1}]$.
We then compare these approaches with the computed FFA values, which
are also in $[\text{0}, \text{1}]$.
The \emph{error} is measured as Manhattan distance,
i.e. the sum of absolute differences across all features.
%
%
We also compare feature rankings according to the competitors
(again using absolute values for LIME and SHAP) using Kendall's Tau~\cite{kendall1938}
and rank-biased overlap~(RBO)~\cite{wmz-tois10} metrics.\footnote{Kendall's
Tau is a correlation coefficient
assessing the ordinal association
between two ranked lists,
offering a measure of similarity in the order of values;
on the other hand, RBO is a metric that measures the
similarity between two ranked lists, taking into account
both the order and the depth of the overlap.}
Kendall's Tau and RBO are measured on a scale $[-\text{1}, \text{1}]$ and $[\text{0}, \text{1}]$, respectively.
A higher value in both metrics
indicates better agreement or closeness
between a ranking and FFA.

\paragraph{Tabular Data.}
\autoref{fig:compas} exemplifies a comparison of FFA, LIME and SHAP on
an instance of the Compas dataset~\cite{propublica16}.
While FFA and LIME agree on the most important feature, ``Asian'',
SHAP gives it very little weight.
Neither LIME nor SHAP agree with FFA, though there is clearly some
similarity.

\autoref{tab:tabffa} details the comparison conducted on 11 tabular
datasets, including \emph{adult}, \emph{compas}, and \emph{recidivism}
datasets commonly used in XAI.
For each dataset, we calculate the metric for each individual instance
and then average the outcomes to obtain the final result for that
dataset.
As can be observed, the errors of LIME's feature attribution across
these datasets span from 1.39 to 5.13.
SHAP demonstrates similar errors within a range $[\text{1.40}, \text{4.76}]$.
LIME and SHAP also exhibit comparable performance in relation to the
two ranking comparison metrics.
The values of Kendall's Tau for LIME (resp. SHAP) are between $-$0.36 and
0.22 (resp. $-$0.39 and 0.27).
Regarding the RBO values, LIME exhibits values between 0.39 and 0.68,
whereas SHAP demonstrates values ranging from 0.44 to 0.67.
Overall, as \autoref{tab:tabffa} indicates, both LIME and SHAP fail
to get close enough to FFA.

\begin{table}[t!]
\caption{Comparison on $\text{10}\times\text{10}$ Images of FFA versus LIME, SHAP and
  FFA approximations.}
  \label{tab:10res}
\centering
\scalebox{0.77}{
	\begin{tabular}{ccccccccc}  \toprule
\textbf{Dataset} & \textbf{LIME} & \textbf{SHAP} & \textbf{\ffalabel{10}} & \textbf{\ffalabel{30}} & \textbf{\ffalabel{60}} & \textbf{\ffalabel{120}} & \textbf{\ffalabel{600}} & \textbf{\ffalabel{1200}}  \\ \cmidrule(lr){2-9}
$\left(|\fml{F}|=\text{100}\right)$ & \multicolumn{8}{c}{\textbf{Error}}  \\ \midrule
10$\times$10-mnist-1vs3 & 11.50 & 10.07 & 5.74 & 5.33 & 4.97 & 4.62 & 3.37 & 2.67  \\
10$\times$10-mnist-1vs7 & 12.64 & 8.28 & 4.16 & 3.58 & 2.94 & 2.50 & 1.42 & 1.01  \\
10$\times$10-pneumoniamnist & 17.32 & 17.90 & 5.37 & 4.32 & 3.78 & 3.39 & 2.22 & 1.64  \\  \midrule
 & \multicolumn{8}{c}{\textbf{Kendall's Tau}}  \\ \midrule
10$\times$10-mnist-1vs3 & -0.15 & 0.48 & 0.49 & 0.57 & 0.62 & 0.65 & 0.74 & 0.80  \\
10$\times$10-mnist-1vs7 & -0.33 & 0.47 & 0.52 & 0.63 & 0.70 & 0.77 & 0.85 & 0.89  \\
10$\times$10-pneumoniamnist & -0.02 & 0.24 & 0.58 & 0.71 & 0.79 & 0.80 & 0.89 & 0.92  \\  \midrule
 & \multicolumn{8}{c}{\textbf{RBO}}  \\ \midrule
10$\times$10-mnist-1vs3 & 0.20 & 0.50 & 0.61 & 0.65 & 0.69 & 0.74 & 0.81 & 0.84  \\
10$\times$10-mnist-1vs7 & 0.19 & 0.58 & 0.73 & 0.77 & 0.81 & 0.86 & 0.90 & 0.90  \\
10$\times$10-pneumoniamnist & 0.21 & 0.37 & 0.61 & 0.70 & 0.73 & 0.77 & 0.83 & 0.87  \\  \bottomrule
	\end{tabular}
}
\end{table}

\paragraph{$\text{10}\times\text{10}$ Digits.}
We now compare the results on $\text{10}\times\text{10}$ downscaled
MNIST digits and PneumoniaMNIST images, where it is feasible to
compute all AXp's.
\autoref{tab:10res} compares LIME's, SHAP's feature attribution and
approximate FFA.
Here, we run AXp enumeration for a number of seconds, which is denoted
as FFA$_{\ast}$, ${\ast} \in \mathbb{R}^+$.
The runtime required for each image by LIME and SHAP is less than one
second.
The results show that the errors of our approximation are small, even
after 10 seconds it beats both LIME and SHAP, and decreases as we
generate more AXp's.
The results for the orderings show again that after 10 seconds,
\ffalabel{$*$} ordering gets closer to the exact FFA than both LIME
and SHAP. Observe how LIME is particularly far away from the
\emph{exact} FFA ordering.

\begin{summary*}
  These results make us confident that we can get useful
  approximations to the exact FFA without exhaustively computing all
  AXp's while feature attribution determined by LIME and SHAP is quite
  erroneous and fails to provide a human-decision maker with useful
  insights, despite being fast.
\end{summary*}


\begin{figure}[t]
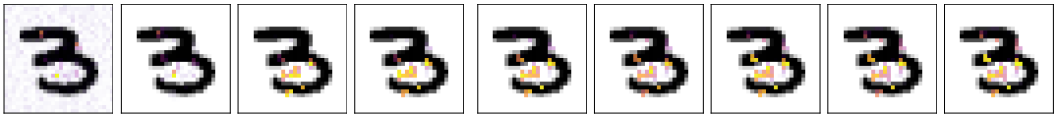

  \centering
   \begin{subfigure}[b]{0.105\textwidth}
    \centering
    \includegraphics[width=\textwidth]{28,28_mnist_1v3_inst13_lime_ori}
    \caption{LIME}
    \label{fig:2828-1v3-lime}
   \end{subfigure}%
   \hspace{0.01cm}
   \begin{subfigure}[b]{0.105\textwidth}
    \centering
    \includegraphics[width=\textwidth]{28,28_mnist_1v3_inst13_shap_ori}
    \caption{SHAP}
    \label{fig:2828-1v3-shap}
   \end{subfigure}%
   \hspace{0.001cm}
   \begin{subfigure}[b]{0.105\textwidth}
    \centering
    \includegraphics[width=\textwidth]{28,28_mnist_1v3_inst13_time10_ori}
    \caption{\ffalabel{10}}
    \label{fig:2828-1v3-10s}
   \end{subfigure}%
   \hspace{0.001cm}
   \begin{subfigure}[b]{0.105\textwidth}
    \centering
    \includegraphics[width=\textwidth]{28,28_mnist_1v3_inst13_time30_ori}
    \caption{\ffalabel{30}}
    \label{fig:2828-1v3-30s}
   \end{subfigure}%
   \hspace{0.001cm}
   \hspace{0.001cm}
   \begin{subfigure}[b]{0.105\textwidth}
    \centering
    \includegraphics[width=\textwidth]{28,28_mnist_1v3_inst13_time120_ori}
    \caption{\ffalabel{120}}
    \label{fig:2828-1v3-120s}
   \end{subfigure}%
   \hspace{0.001cm}
   \begin{subfigure}[b]{0.105\textwidth}
    \centering
    \includegraphics[width=\textwidth]{28,28_mnist_1v3_inst13_time600_ori}
    \caption{\ffalabel{600}}
    \label{fig:2828-1v3-600s}
   \end{subfigure}%
   \hspace{0.001cm}
   \begin{subfigure}[b]{0.105\textwidth}
    \centering
    \includegraphics[width=\textwidth]{28,28_mnist_1v3_inst13_time1200_ori}
    \caption{\ffalabel{1.2k}}
    \label{fig:2828-1v3-1200s}
   \end{subfigure}%
   \hspace{0.001cm}
   \begin{subfigure}[b]{0.105\textwidth}
    \centering
    \includegraphics[width=\textwidth]{28,28_mnist_1v3_inst13_time3600_ori}
    \caption{\ffalabel{3.6k}}
    \label{fig:2828-1v3-3600s}
   \end{subfigure}%
   \hspace{0.001cm}
   \begin{subfigure}[b]{0.105\textwidth}
    \centering
    \includegraphics[width=\textwidth]{28,28_mnist_1v3_inst13_time7200_ori}
    \caption{\ffalabel{7.2k}}
    \label{fig:2828-1v3-7200s}
   \end{subfigure}%
   \caption{28 $\times$ 28 MNIST 1 vs. 3. The prediction is digit 3. The \href{https://matplotlib.org/stable/tutorials/colors/colormaps.html}{\emph{plasma} gradient} is used ranging from deep purple for the least important features to vibrant yellow for the most important features.
}
    \label{fig:2828-1v3}
\end{figure}

\begin{figure}[t]
  \centering
   \begin{subfigure}[b]{0.105\textwidth}
    \centering
    \includegraphics[width=\textwidth]{28,28_mnist_1v7_inst4_lime_ori}
    \caption{LIME}
    \label{fig:2828-1v7-lime}
   \end{subfigure}%
   \hspace{0.01cm}
   \begin{subfigure}[b]{0.105\textwidth}
    \centering
    \includegraphics[width=\textwidth]{28,28_mnist_1v7_inst4_shap_ori}
    \caption{SHAP}
    \label{fig:2828-1v7-shap}
   \end{subfigure}%
   \hspace{0.01cm}
   \begin{subfigure}[b]{0.105\textwidth}
    \centering
    \includegraphics[width=\textwidth]{28,28_mnist_1v7_inst4_time10_ori}
    \caption{\ffalabel{10}}
    \label{fig:2828-1v7-10s}
   \end{subfigure}%
   \hspace{0.01cm}
   \begin{subfigure}[b]{0.105\textwidth}
    \centering
    \includegraphics[width=\textwidth]{28,28_mnist_1v7_inst4_time30_ori}
    \caption{\ffalabel{30}}
    \label{fig:2828-1v7-30s}
   \end{subfigure}%
   \hspace{0.01cm}
   \begin{subfigure}[b]{0.105\textwidth}
    \centering
    \includegraphics[width=\textwidth]{28,28_mnist_1v7_inst4_time120_ori}
    \caption{\ffalabel{120}}
    \label{fig:2828-1v7-120s}
   \end{subfigure}%
   \hspace{0.01cm}
   \begin{subfigure}[b]{0.105\textwidth}
    \centering
    \includegraphics[width=\textwidth]{28,28_mnist_1v7_inst4_time600_ori}
    \caption{\ffalabel{600}}
    \label{fig:2828-1v7-600s}
   \end{subfigure}%
   \hspace{0.01cm}
   \begin{subfigure}[b]{0.105\textwidth}
    \centering
    \includegraphics[width=\textwidth]{28,28_mnist_1v7_inst4_time1200_ori}
    \caption{\ffalabel{1.2k}}
    \label{fig:2828-1v7-1200s}
   \end{subfigure}%
   \hspace{0.01cm}
   \begin{subfigure}[b]{0.105\textwidth}
    \centering
    \includegraphics[width=\textwidth]{28,28_mnist_1v7_inst4_time3600_ori}
    \caption{\ffalabel{3.6k}}
    \label{fig:2828-1v7-3600s}
   \end{subfigure}%
   \hspace{0.01cm}
   \begin{subfigure}[b]{0.105\textwidth}
    \centering
    \includegraphics[width=\textwidth]{28,28_mnist_1v7_inst4_time7200_ori}
    \caption{\ffalabel{7.2k}}
    \label{fig:2828-1v7-7200s}
   \end{subfigure}%
     \caption{28 $\times$ 28 MNIST 1 vs. 7. The prediction is digit 7.}
    \label{fig:2828-1v7}
\end{figure}

\begin{figure}[t]
  \centering
   \begin{subfigure}[b]{0.105\textwidth}
    \centering
    \includegraphics[width=\textwidth]{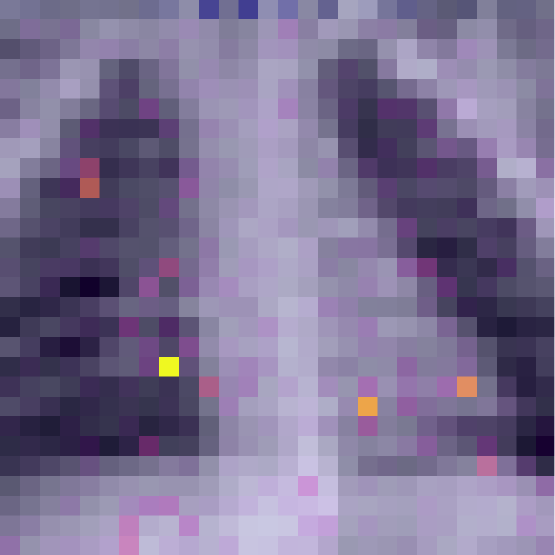}
    \caption{LIME}
    \label{fig:2828-pneumonia-lime}
   \end{subfigure}%
   \hspace{0.01cm}
   \begin{subfigure}[b]{0.105\textwidth}
    \centering
    \includegraphics[width=\textwidth]{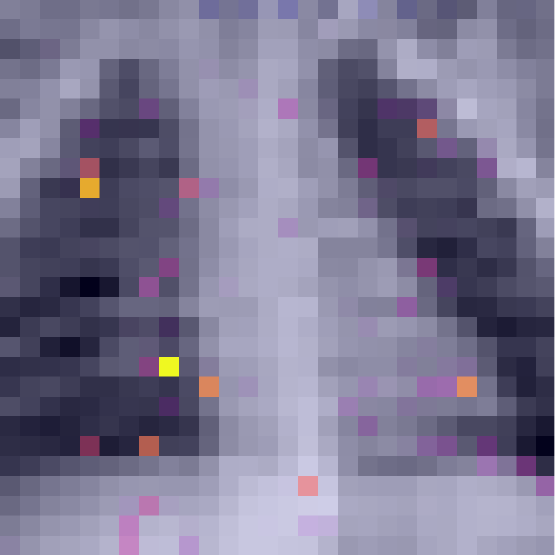}
    \caption{SHAP}
    \label{fig:2828-pneumonia-shap}
   \end{subfigure}%
    \hspace{0.01cm}
   \begin{subfigure}[b]{0.105\textwidth}
    \centering
    \includegraphics[width=\textwidth]{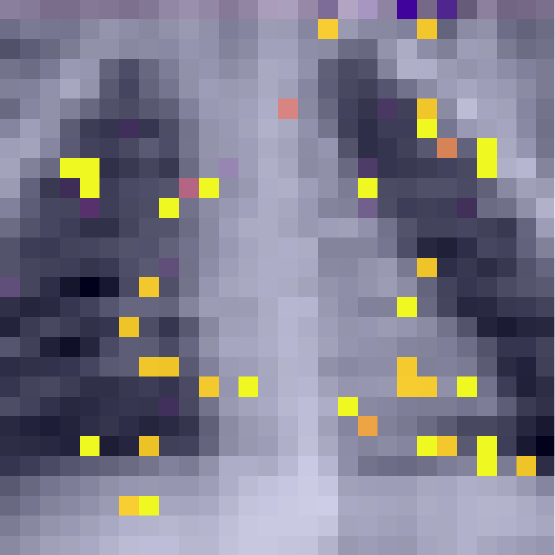}
    \caption{\ffalabel{10}}
    \label{fig:2828-pneumonia-10s}
   \end{subfigure}%
   \hspace{0.01cm}
   \begin{subfigure}[b]{0.105\textwidth}
    \centering
    \includegraphics[width=\textwidth]{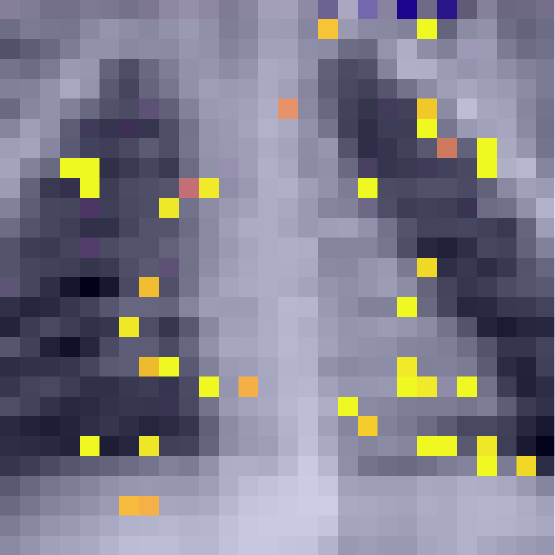}
    \caption{\ffalabel{30}}
    \label{fig:2828-pneumonia-30s}
   \end{subfigure}%
   \hspace{0.01cm}
   \begin{subfigure}[b]{0.105\textwidth}
    \centering
    \includegraphics[width=\textwidth]{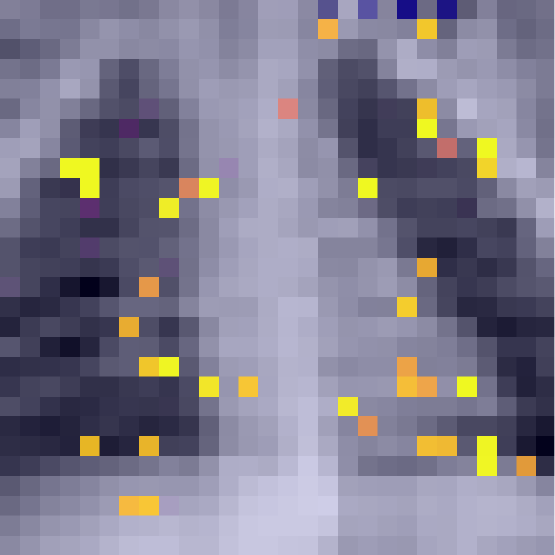}
    \caption{\ffalabel{120}}
    \label{fig:2828-pneumonia-120s}
   \end{subfigure}%
   \hspace{0.01cm}
   \begin{subfigure}[b]{0.105\textwidth}
    \centering
    \includegraphics[width=\textwidth]{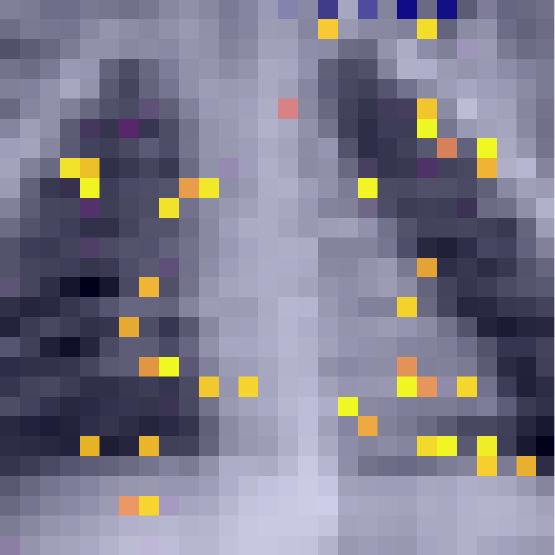}
    \caption{\ffalabel{600}}
    \label{fig:2828-pneumonia-600s}
   \end{subfigure}%
   \hspace{0.01cm}
   \begin{subfigure}[b]{0.105\textwidth}
    \centering
    \includegraphics[width=\textwidth]{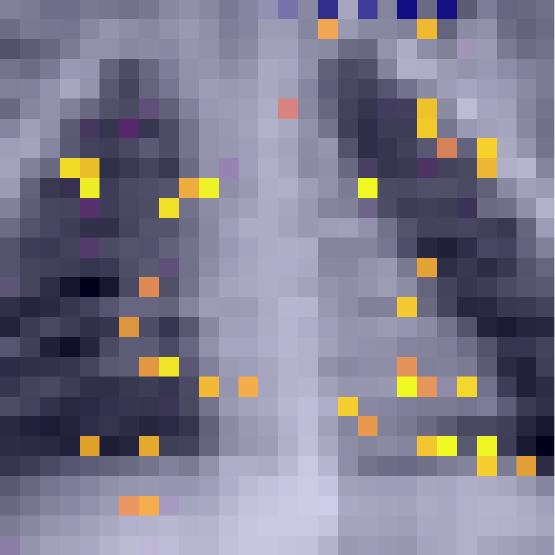}
    \caption{\ffalabel{1.2k}}
    \label{fig:2828-pneumonia-1200s}
   \end{subfigure}%
   \hspace{0.01cm}
   \begin{subfigure}[b]{0.105\textwidth}
    \centering
    \includegraphics[width=\textwidth]{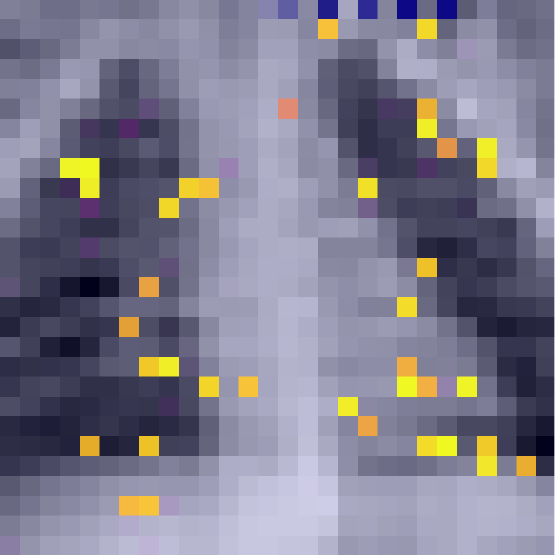}
    \caption{\ffalabel{3.6k}}
    \label{fig:2828-pneumonia-3600s}
   \end{subfigure}%
   \hspace{0.01cm}
   \begin{subfigure}[b]{0.105\textwidth}
    \centering
    \includegraphics[width=\textwidth]{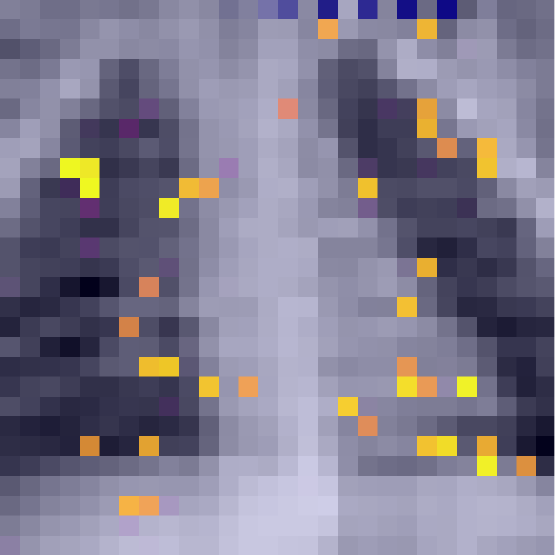}
    \caption{\ffalabel{7.2k}}
    \label{fig:2828-pneumonia-7200s}
   \end{subfigure}%
     \caption{28 $\times$ 28 PneumoniaMNIST. The prediction is normal.}
    \label{fig:2828-pneumonia}
\end{figure}

\begin{table}[t!]
 \caption{Comparison on 28 $\times$ 28 Images of \ffalabel{7200} versus LIME, SHAP and
  FFA approximations.}
\label{tab:28res}
\centering
\scalebox{0.77}{
	\begin{tabular}{ccccccccc}  \toprule
	  \textbf{Dataset} & \textbf{LIME} & \textbf{SHAP} & \textbf{\ffalabel{10}} & \textbf{\ffalabel{30}} & \textbf{\ffalabel{120}} & \textbf{\ffalabel{600}} & \textbf{\ffalabel{1200}} & \textbf{\ffalabel{3600}}  \\ \cmidrule(lr){2-9}
$\left(|\fml{F}|=\text{784}\right)$ & \multicolumn{8}{c}{\textbf{Error}}  \\ \midrule
28$\times$28-mnist-1vs3 & 49.66 & 22.77 & 9.44 & 7.61 & 6.81 & 4.51 & 3.13 & 2.69  \\
28$\times$28-mnist-1vs7 & 55.10 & 24.92 & 11.78 & 9.58 & 6.94 & 4.51 & 3.30 & 2.18  \\
28$\times$28-pneumoniamnist & 62.94 & 31.55 & 8.17 & 7.81 & 5.69 & 4.89 & 3.77 & 3.10  \\  \midrule
 & \multicolumn{8}{c}{\textbf{Kendall's Tau}}  \\ \midrule
28$\times$28-mnist-1vs3 & -0.80 & 0.42 & 0.44 & 0.62 & 0.69 & 0.80 & 0.86 & 0.87  \\
28$\times$28-mnist-1vs7 & -0.79 & 0.34 & 0.40 & 0.56 & 0.72 & 0.82 & 0.87 & 0.92  \\
28$\times$28-pneumoniamnist & -0.66 & 0.24 & 0.34 & 0.50 & 0.67 & 0.76 & 0.80 & 0.87  \\  \midrule
 & \multicolumn{8}{c}{\textbf{RBO}}  \\ \midrule
28$\times$28-mnist-1vs3 & 0.03 & 0.40 & 0.43 & 0.50 & 0.61 & 0.78 & 0.83 & 0.88  \\
28$\times$28-mnist-1vs7 & 0.03 & 0.34 & 0.40 & 0.45 & 0.58 & 0.76 & 0.83 & 0.93  \\
28$\times$28-pneumoniamnist & 0.03 & 0.23 & 0.31 & 0.35 & 0.42 & 0.59 & 0.66 & 0.83  \\  \bottomrule
	\end{tabular}
}
\end{table}

\subsection{Approximating Formal Feature Attribution}

Since the problem of formal feature attribution ``lives'' in
$\Sigma_2^\text{P}$, it is not surprising that computing FFA may be
challenging in practice.
\autoref{tab:10res} suggests that our approach gets good FFA
approximations even if we only collect AXp's for a short time.
Here we compare the fidelity of our approach versus the approximate
FFA computed after 2 hours~(7200s).
\autoref{fig:2828-1v3}, \ref{fig:2828-1v7}, and
\ref{fig:2828-pneumonia} depict feature attributions generated by
LIME, SHAP and FFA$_\ast$ for the three selected 28 $\times$ 28
images.
The comparison between LIME, SHAP, and the approximate FFA computation
is detailed in \autoref{tab:28res}.
The LIME and SHAP processing time for each image is less than one
second.
The average findings detailed in \autoref{tab:28res} are consistent
with those shown in \autoref{tab:10res}.
Namely, FFA approximation yields better errors, Kendall’s Tau and RBO
values, outperforming both LIME, and SHAP after 10 seconds.
Furthermore, the results demonstrate that after 10 seconds our
approach places feature attributions closer to \ffalabel{7200}
compared to both LIME and SHAP hinting on the features that are truly
relevant for the prediction.


\begin{table}[t!]
\caption{Just-in-Time Defect Prediction comparison of FFA versus LIME and SHAP.}
\label{tab:jit}
\centering
\scalebox{0.77}{
	\begin{tabular}{ccccccc}  \toprule
	  \multirow{2}{*}{\textbf{Approach}} & \multicolumn{3}{c}{\textbf{openstack}~\small$\left(|\fml{F}|=\text{13}\right)$} & \multicolumn{3}{c}{\textbf{qt}~\small$\left(|\fml{F}|=\text{16}\right)$}  \\ \cmidrule{2-7}
  & \textbf{Error} & \textbf{Kendall's Tau} & \textbf{RBO} & \textbf{Error} & \textbf{Kendall's Tau} & \textbf{RBO}  \\ \midrule
LIME & 4.84 & 0.05 & 0.55 & 5.63 & -0.08 & 0.45  \\
SHAP & 5.08 & 0.00 & 0.53 & 5.22 & -0.13 & 0.44  \\  \bottomrule
	\end{tabular}
}
\end{table}

\subsection{Application in Just-in-Time Defect Prediction}
Just-in-Time (JIT) defect
prediction~\cite{kim2007predicting,Kamei2013,pornprasit2021jitline,lin2021impact}
has been recently proposed to predict if a commit will introduce
software defects in the future, enabling development teams to
prioritize their limited Software Quality Assurance resources on the
most risky commits/pull requests.
The approach of JIT defect prediction has often been considered a
black-box, lacking explainability for practitioners.
To tackle this challenge, our proposed approach to generating FFA can
be employed, as model-agnostic approaches cannot guarantee to provide
accurate feature attribution (see above).
We use logistic regression models of~\cite{pornprasit2021pyexplainer}
based on large-scale open-source Openstack and Qt datasets provided
by~\cite{mcintosh2017fix} commonly used for JIT defect
prediction~\cite{pornprasit2021pyexplainer}.
Monotonicity of logistic regression enables us to enumerate
explanations using the approach of~\cite{msgcin-icml21} and so to
extract \emph{exact FFA} for each instance \emph{within a second}.
\autoref{tab:jit} details the comparison of FFA, LIME and SHAP in
terms of the three considered metrics.
As with the outcomes presented in \autoref{tab:tabffa},
\autoref{tab:10res}, and \autoref{tab:28res}, neither LIME nor SHAP
align with formal feature attribution, though there are some
similarities between them.

\section{Limitations} \label{sec:limits}


Despite the rigorous guarantees provided by formal feature attribution
and high-quality of the result explanations, the following limitations
can be identified.
First, our approach relies on formal reasoning and thus requires an ML
model of interest to admit a representation in some fragments of
first-order logic, and the corresponding reasoner to deal with
it~\cite{msi-aaai22}.
Second, the problem complexity impedes immediate and widespread use of
FFA and signifies the need to develop effective methods of FFA
approximation.
Finally, though our experimental evidence suggests that FFA
approximations quickly converge to the exact values of FFA, whether or
not this holds in general remains an open question.

\section{Conclusions} \label{sec:conc}

Most approaches to XAI are heuristic methods that are susceptible to
unsoundness and out-of-distribution sampling.
Formal approaches to XAI have so far concentrated on the problem of
feature selection, detecting which features are important for
justifying a classification decision, and not on feature attribution,
where we can understand the weight of a feature in making such a
decision.
In this paper we define the first formal approach to feature
attribution (FFA) we are aware of, using the proportion of abductive
explanations in which a feature occurs to weight its importance.
We show that we can compute FFA exactly for many classification
problems, and when we cannot we can compute effective approximations.
Existing heuristic approaches to feature attribution do not agree with
FFA.
Sometimes they markedly differ, for example, assigning no weight to a
feature that appears in (a large number of) explanations, or assigning
(large) non-zero weight to a feature that is irrelevant for the
prediction.
Overall, the paper argues that if we agree that FFA is a correct
measure of feature attribution then we need to investigate methods
that compute good FFA approximations quickly.

\clearpage  
\bibliographystyle{abbrvnat}
\bibliography{refs}

\newpage
\clearpage
\appendix
\appendixpage


\section{Exact Weighted Formal Feature Attribution}\label{sec:wffares}

In this appendix, we once again limit our analysis to instances where
we can calculate the \emph{exact} WFFA values for the instance of
interest by enumerating all AXp’s.
Also, the settings used in \autoref{sec:res} are applied here, i.e. we
take the absolute values of feature attibution assignned by LIME and
SHAP, and normalize them within the range of $[\text{0}, \text{1}]$.
Just like in the main text of the paper, we then compare these
approaches with normalized WFFA values in terms of errors, Kendall’s
Tau~\cite{kendall1938} and rank-biased
overlap~(RBO)~\cite{wmz-tois10}.

\subsection{Tabular Data}
\begin{figure}[t]
  \centering
   \begin{subfigure}[b]{0.33\textwidth}
    \centering
    \includegraphics[width=\textwidth]{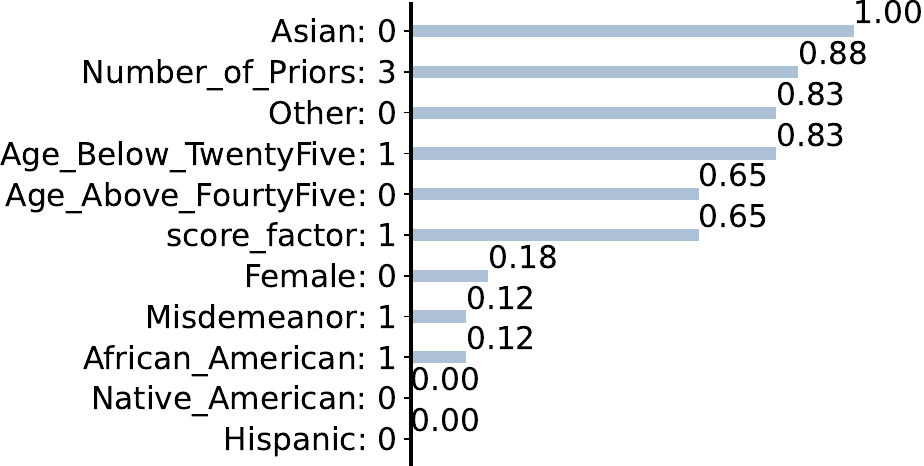}
    \caption{WFFA}
    \label{fig:formal-compas-wffa}
   \end{subfigure}%
   \begin{subfigure}[b]{0.33\textwidth}
    \centering
    \includegraphics[width=\textwidth]{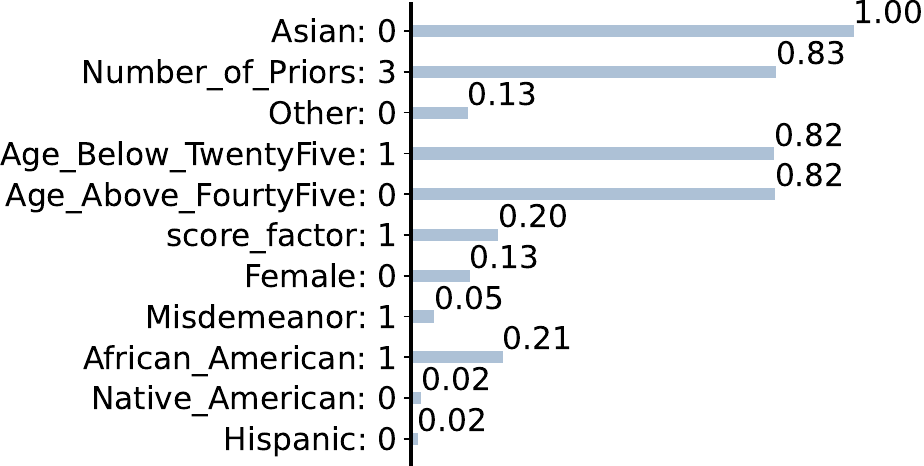}
    \caption{LIME}
    \label{fig:lime-compas-wffa}
   \end{subfigure}%
   \begin{subfigure}[b]{0.33\textwidth}
    \centering
    \includegraphics[width=\textwidth]{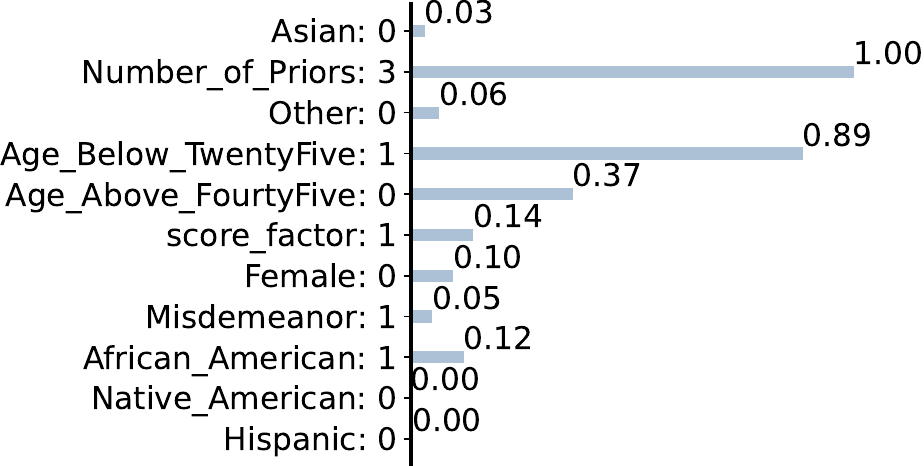}
    \caption{SHAP}
    \label{fig:shap-compas-wffa}
   \end{subfigure}%
    \caption{Explanations for an instance of Compas
		$\mbf{v} =  \{
			\text{\#Priors}=\text{3}, \text{Score\_factor}=\text{1},
			\text{Age\_Above\_FourtyFive}=\text{0}, \text{Age\_Below\_TwentyFive}=\text{1},
 			\text{African\_American}=\text{1}, \text{Asian}=\text{0},
			\text{Hispanic}=\text{0}, \text{Native\_American}=\text{0},
			\text{Other}=\text{0}, \text{Female}=\text{0},
			\text{Misdemeanor}=\text{1}
			\}$
	    predicted as Two\_yr\_Recidivism $=$ true.
    }
    \label{fig:compas-wffa}
\end{figure}

A comparison of WFFA, LIME and SHAP on an instance of the Compas
dataset~\cite{propublica16} is exemplified in
\autoref{fig:compas-wffa}.
We can observe the patterns similar to those depicted in
\autoref{fig:compas}.
The feature that WFFA considers most important is ``Asian'' while this
viewpoint is shared by LIME but disputed by SHAP.
However, neither LIME nor SHAP fully align with WFFA, although there
is evident similarity between them.
As with FFA, these observations can be generalized to the other
instances of Compas, as discussed below.

\autoref{tab:tabwffa} presents a comparison of WFFA against LIME, and
SHAP on the 11 selected tabular datasets as in \autoref{tab:tabffa},
demonstrating similarities in the findings observed for WFFA and FFA
for these datasets.
The average runtime for generating the exact WFFA in a dataset varies
between 0.18 and 1.89 seconds while the average number of AXp's per
instance to explain and so to compute exact WFFA in a dataset ranges
from 1.40 to 33.33.
Both LIME and SHAP process each image in less than one second.
%
LIME exhibits errors ranging from 1.37 to 4.96 across these datasets
while SHAP shows similar errors spanning from 1.36 to 4.67.
Besides errors, LIME and SHAP yield comparable outcomes in terms of
the two ranking comparison metrics.
The values of Kendall's Tau for LIME span from $-$0.35 to 0.25,
whereas the values for SHAP are between $-$0.38 and 0.31.
Regarding RBO values, LIME (resp. SHAP) demonstrates values ranging
from 0.38 to 0.69 (resp. 0.43 to 0.67).
Overall and consistent with the FFA findings shown earlier in
\autoref{tab:tabffa}, \autoref{tab:tabwffa} indicates that both LIME
and SHAP fail to achieve close enough agreement with WFFA.

\subsection{10 $\times$ 10 Digits}

\begin{figure}[t]
  \centering
   \begin{subfigure}[b]{0.09\textwidth}
    \centering
    \includegraphics[width=\textwidth]{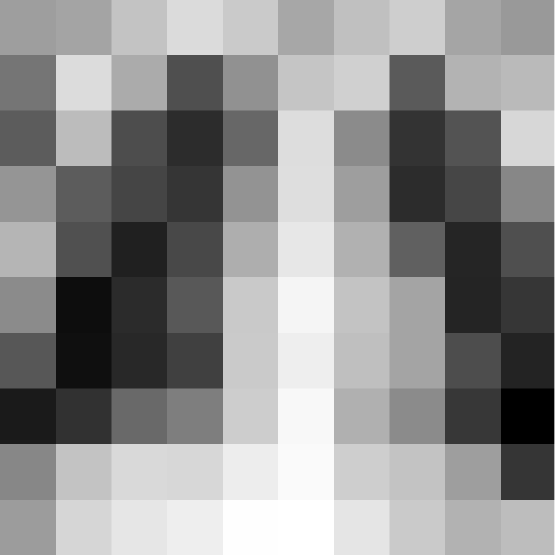}
    {\small Image}
    \label{fig:1010-pneumonia-input}
   \end{subfigure}%
   \hspace{0.01cm}
   \begin{subfigure}[b]{0.09\textwidth}
    \centering
    {\includegraphics[width=\textwidth]{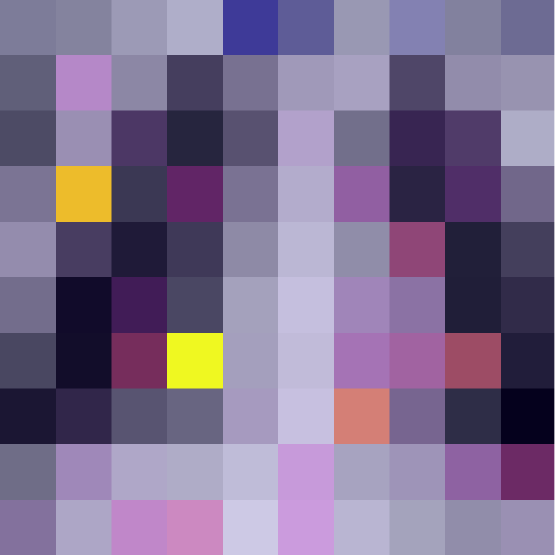}}
    {\small LIME}
    \label{fig:1010-pneumonia-lime}
   \end{subfigure}%
   \hspace{0.01cm}
   \begin{subfigure}[b]{0.09\textwidth}
    \centering
    {\includegraphics[width=\textwidth]{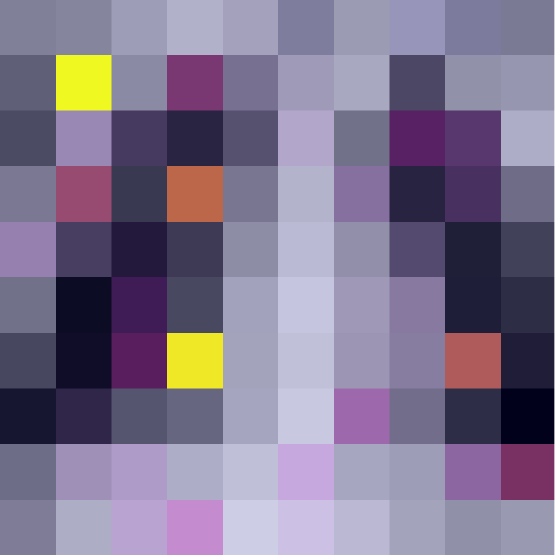}}
    {\small SHAP}
    \label{fig:1010-pneumonia-shap}
   \end{subfigure}%
   \hspace{0.01cm}
   \begin{subfigure}[b]{0.09\textwidth}
    \centering
    \includegraphics[width=\textwidth]{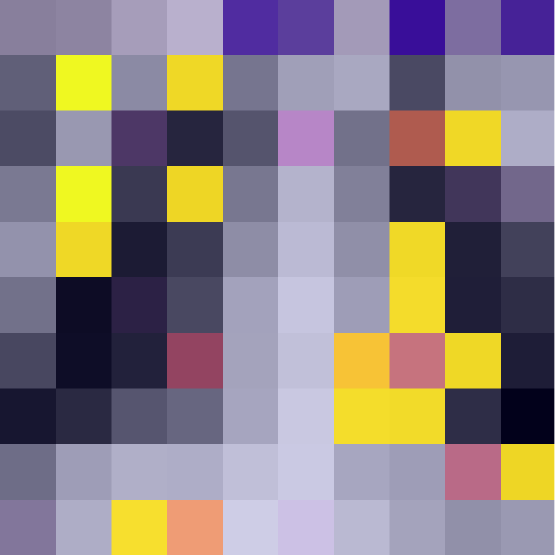}
    {\small \ffalabel{10}}
    \label{fig:1010-pneumonia-10s}
   \end{subfigure}%
   \hspace{0.01cm}
   \begin{subfigure}[b]{0.09\textwidth}
    \centering
    \includegraphics[width=\textwidth]{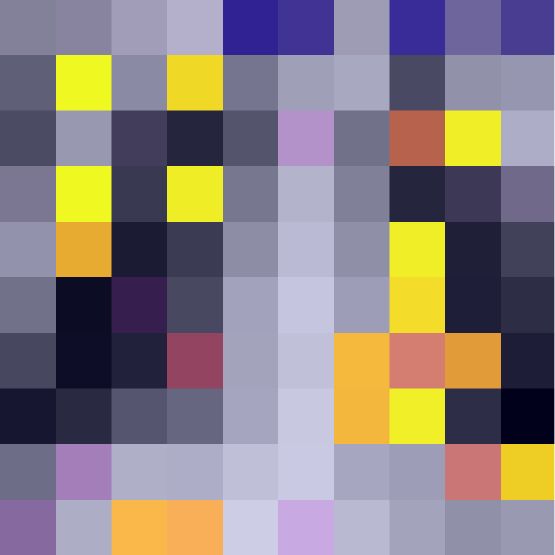}
    {\small \ffalabel{30}}
    \label{fig:1010-pneumonia-30s}
   \end{subfigure}%
   \hspace{0.01cm}
   \begin{subfigure}[b]{0.09\textwidth}
    \centering
    \includegraphics[width=\textwidth]{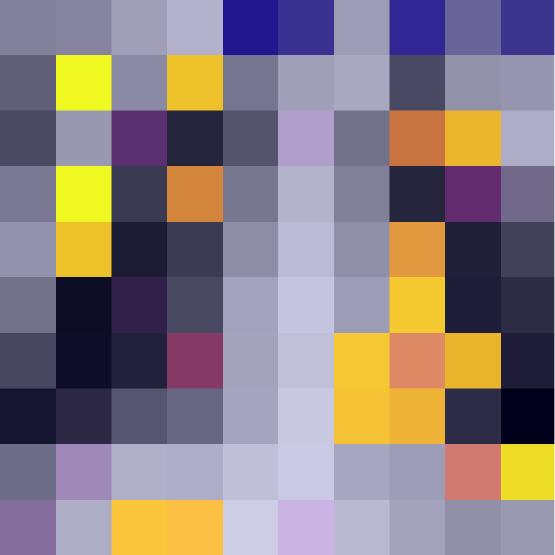}
    {\small \ffalabel{60}}
    \label{fig:1010-pneumonia-60s}
   \end{subfigure}%
   \hspace{0.01cm}
   \begin{subfigure}[b]{0.09\textwidth}
    \centering
    \includegraphics[width=\textwidth]{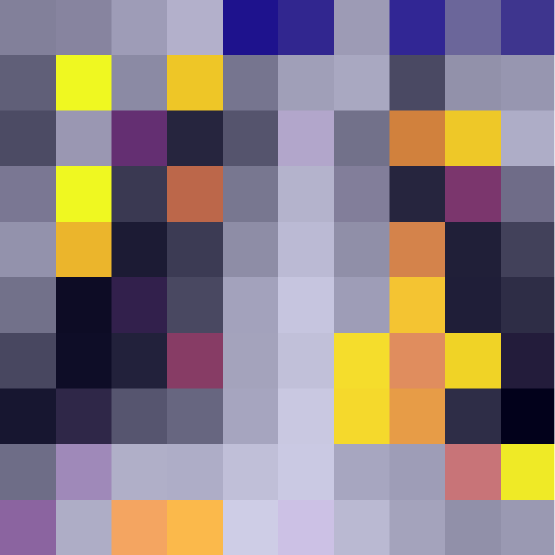}
    {\small \ffalabel{120}}
    \label{fig:1010-pneumonia-120s}
   \end{subfigure}%
   \hspace{0.01cm}
   \begin{subfigure}[b]{0.09\textwidth}
    \centering
    \includegraphics[width=\textwidth]{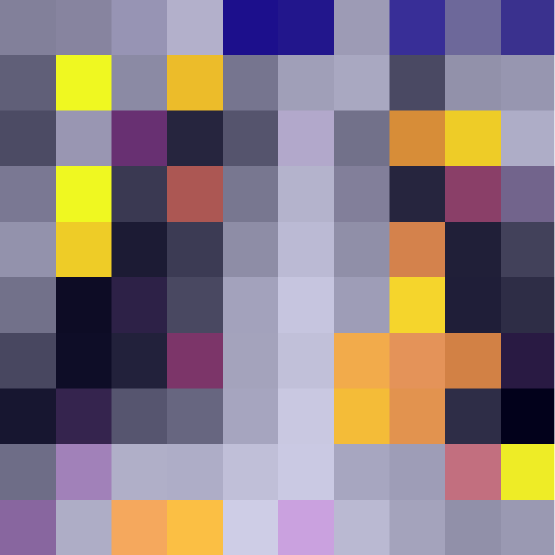}
    {\small \ffalabel{600}}
    \label{fig:1010-pneumonia-600s}
   \end{subfigure}%
   \hspace{0.01cm}
   \begin{subfigure}[b]{0.09\textwidth}
    \centering
    \includegraphics[width=\textwidth]{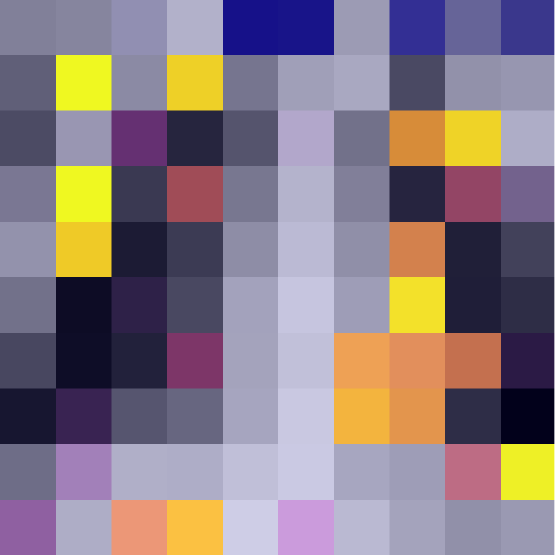}
    {\small \ffalabel{1.2k}}
    \label{fig:1010-pneumonia-1200s}
   \end{subfigure}%
   \hspace{0.01cm}
   \begin{subfigure}[b]{0.09\textwidth}
    \centering
    \includegraphics[width=\textwidth]{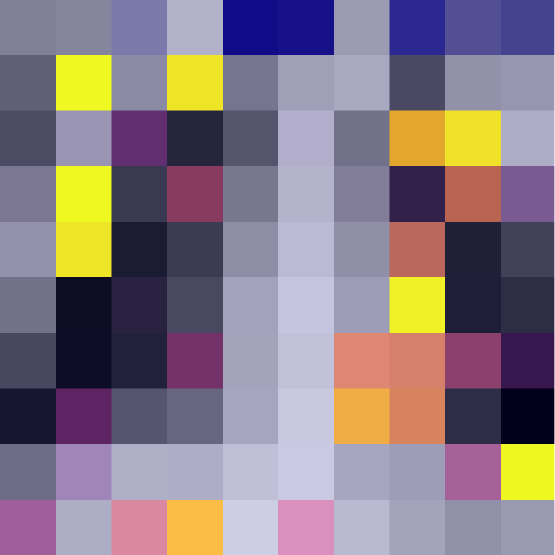}
    {\small FFA}
    \label{fig:1010-pneumonia-attr}
   \end{subfigure}%
     \caption{10 $\times$ 10 PneumoniaMNIST. The prediction is pneumonia.}
    \label{fig:1010-pneumonia}
\end{figure}

\begin{figure}[t]
  \centering
   \begin{subfigure}[b]{0.09\textwidth}
    \centering
    {\includegraphics[width=\textwidth]{10,10_mnist_1v3_inst7_ori}}
    {\small Input}
    \label{fig:1010-1v3-input-wffa}
   \end{subfigure}%
   \hspace{0.01cm}
   \begin{subfigure}[b]{0.09\textwidth}
    \centering
    {\includegraphics[width=\textwidth]{10,10_mnist_1v3_inst7_lime_ori}}
    {\small LIME}
    \label{fig:1010-1v3-lime-wffa}
   \end{subfigure}%
   \hspace{0.01cm}
   \begin{subfigure}[b]{0.09\textwidth}
    \centering
    {\includegraphics[width=\textwidth]{10,10_mnist_1v3_inst7_shap_ori}}
    {\small SHAP}
    \label{fig:1010-1v3-shap-wffa}
   \end{subfigure}%
   \hspace{0.01cm}
   \begin{subfigure}[b]{0.09\textwidth}
    \centering
    {\includegraphics[width=\textwidth]{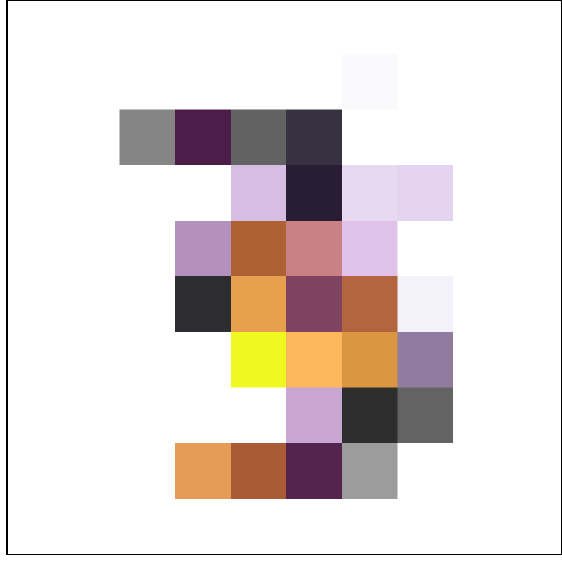}}
    {\small \ffalabel{10}}
    \label{fig:1010-1v3-10s-wffa}
   \end{subfigure}%
   \hspace{0.01cm}
   \begin{subfigure}[b]{0.09\textwidth}
    \centering
    {\includegraphics[width=\textwidth]{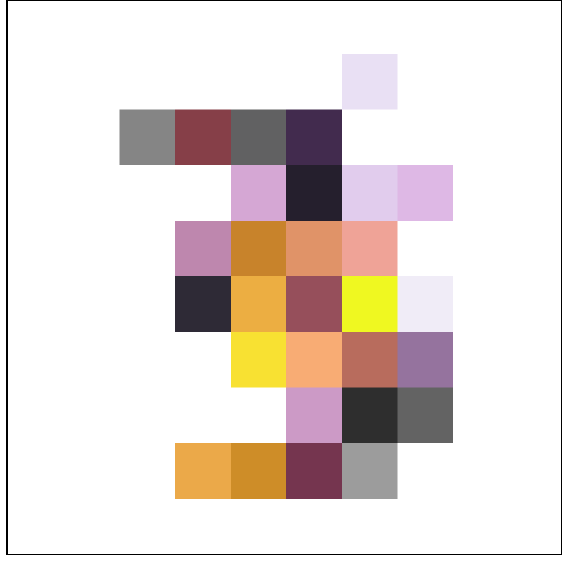}}
    {\small \ffalabel{30}}
    \label{fig:1010-1v3-30s-wffa}
   \end{subfigure}%
   \hspace{0.01cm}
   \begin{subfigure}[b]{0.09\textwidth}
    \centering
    {\includegraphics[width=\textwidth]{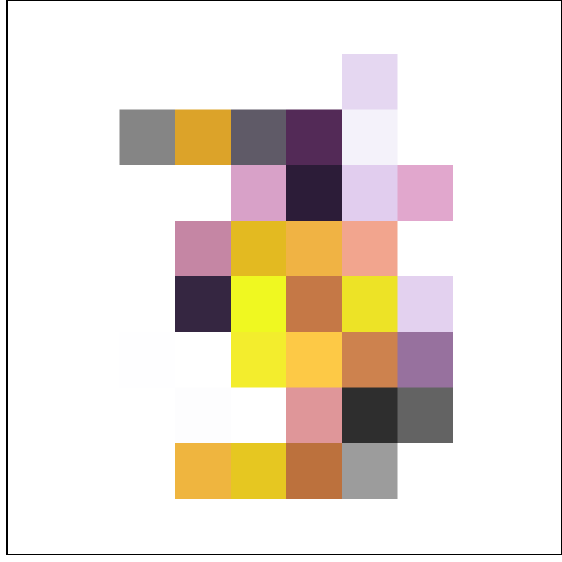}}
    {\small \ffalabel{60}}
    \label{fig:1010-1v3-60s-wffa}
   \end{subfigure}%
   \hspace{0.01cm}
   \begin{subfigure}[b]{0.09\textwidth}
    \centering
    {\includegraphics[width=\textwidth]{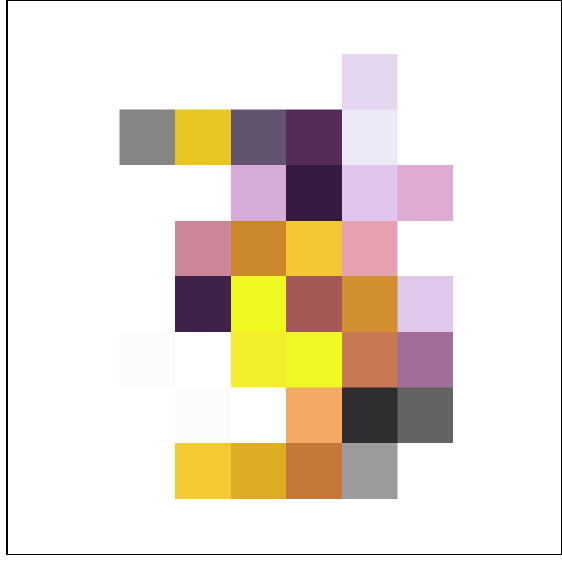}}
    {\small \ffalabel{120}}
    \label{fig:1010-1v3-120s-wffa}
   \end{subfigure}%
   \hspace{0.01cm}
   \begin{subfigure}[b]{0.09\textwidth}
    \centering
    {\includegraphics[width=\textwidth]{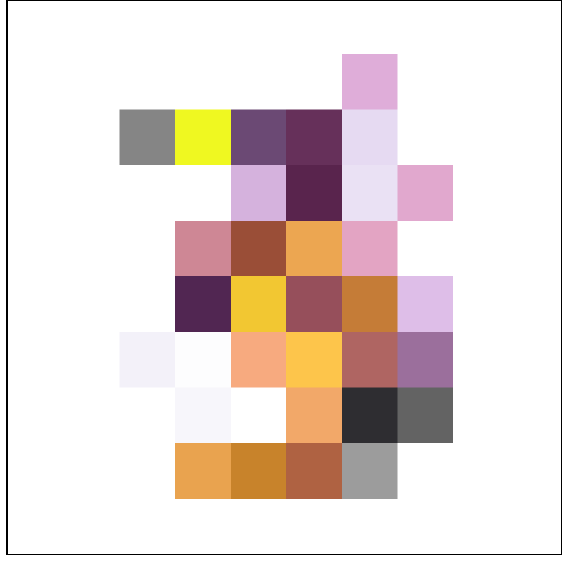}}
    {\small \ffalabel{600}}
    \label{fig:1010-1v3-600s-wffa}
   \end{subfigure}%
   \hspace{0.01cm}
   \begin{subfigure}[b]{0.09\textwidth}
    \centering
    {\includegraphics[width=\textwidth]{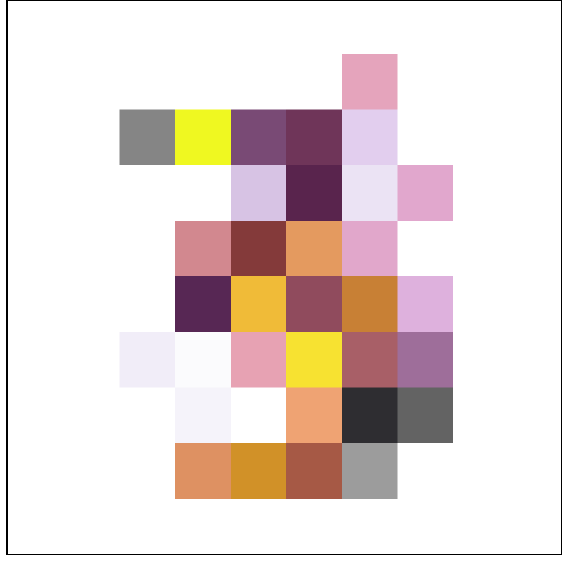}}
    {\small \ffalabel{1.2k}}
    \label{fig:1010-1v3-1200s-wffa}
   \end{subfigure}%
   \hspace{0.01cm}
   \begin{subfigure}[b]{0.09\textwidth}
    \centering
    {\includegraphics[width=\textwidth]{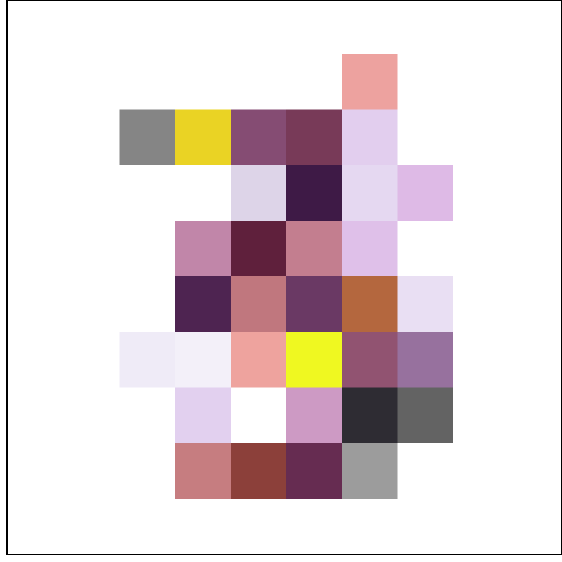}}
    {\small FFA}
    \label{fig:1010-1v3-attr-wffa}
   \end{subfigure}%
     \caption{10 $\times$ 10 MNIST 1 vs. 3. The prediction is 3.}
    \label{fig:1010-1v3-wffa}
\end{figure}

\begin{figure}[t]
  \centering
   \begin{subfigure}[b]{0.09\textwidth}
    \centering
    \includegraphics[width=\textwidth]{10,10_mnist_1v7_inst8_ori}
    {\small Input}
    \label{fig:1010-1v7-input-wffa}
   \end{subfigure}%
   \hspace{0.01cm}
    \begin{subfigure}[b]{0.09\textwidth}
    \centering
    {\includegraphics[width=\textwidth]{10,10_mnist_1v7_inst8_lime_ori}}
    {\small LIME}
    \label{fig:1010-1v7-lime-wffa}
   \end{subfigure}%
   \hspace{0.01cm}
   \begin{subfigure}[b]{0.09\textwidth}
    \centering
    {\includegraphics[width=\textwidth]{10,10_mnist_1v7_inst8_shap_ori}}
    {\small SHAP}
    \label{fig:1010-1v7-shap-wffa}
   \end{subfigure}%
   \hspace{0.01cm}
   \begin{subfigure}[b]{0.09\textwidth}
    \centering
    \includegraphics[width=\textwidth]{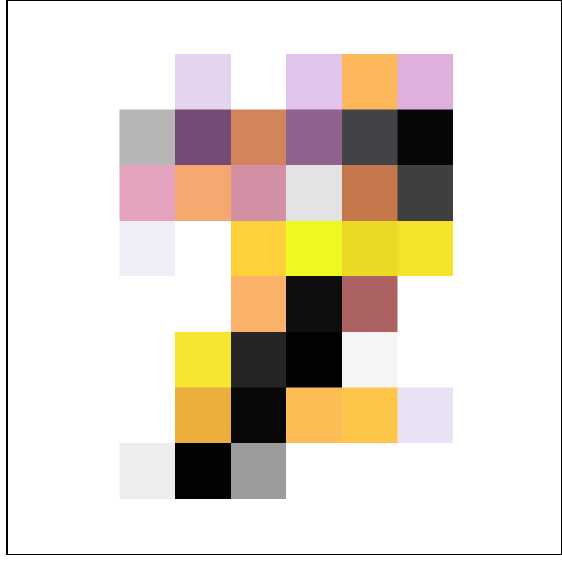}
    {\small \ffalabel{10}}
    \label{fig:1010-1v7-10s-wffa}
   \end{subfigure}%
   \hspace{0.01cm}
   \begin{subfigure}[b]{0.09\textwidth}
    \centering
    \includegraphics[width=\textwidth]{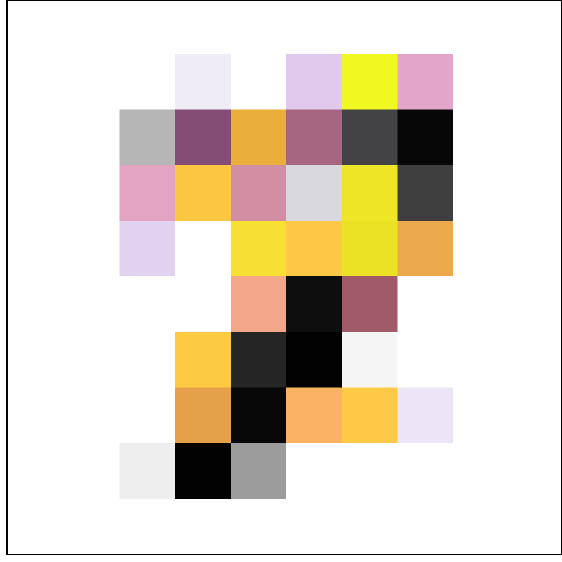}
    {\small \ffalabel{30}}
    \label{fig:1010-1v7-30s-wffa}
   \end{subfigure}%
   \hspace{0.01cm}
   \begin{subfigure}[b]{0.09\textwidth}
    \centering
    \includegraphics[width=\textwidth]{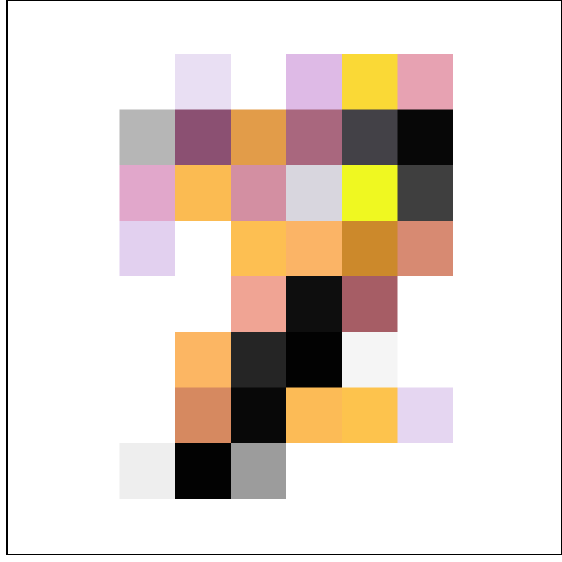}
    {\small \ffalabel{60}}
    \label{fig:1010-1v7-60s-wffa}
   \end{subfigure}%
   \hspace{0.01cm}
   \begin{subfigure}[b]{0.09\textwidth}
    \centering
    \includegraphics[width=\textwidth]{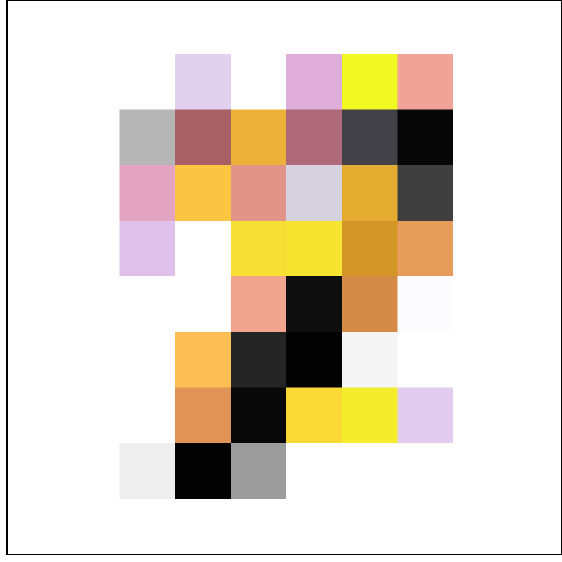}
    {\small \ffalabel{120}}
    \label{fig:1010-1v7-120s-wffa}
   \end{subfigure}%
   \hspace{0.01cm}
   \begin{subfigure}[b]{0.09\textwidth}
    \centering
    \includegraphics[width=\textwidth]{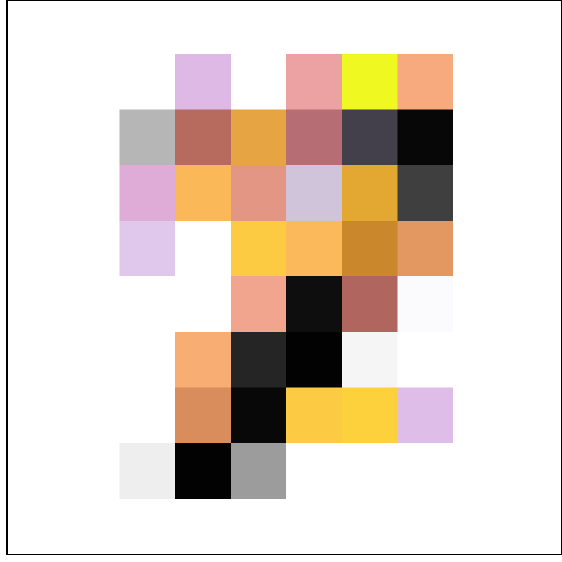}
    {\small \ffalabel{600}}
    \label{fig:1010-1v7-600s-wffa}
   \end{subfigure}%
   \hspace{0.01cm}
   \begin{subfigure}[b]{0.09\textwidth}
    \centering
    \includegraphics[width=\textwidth]{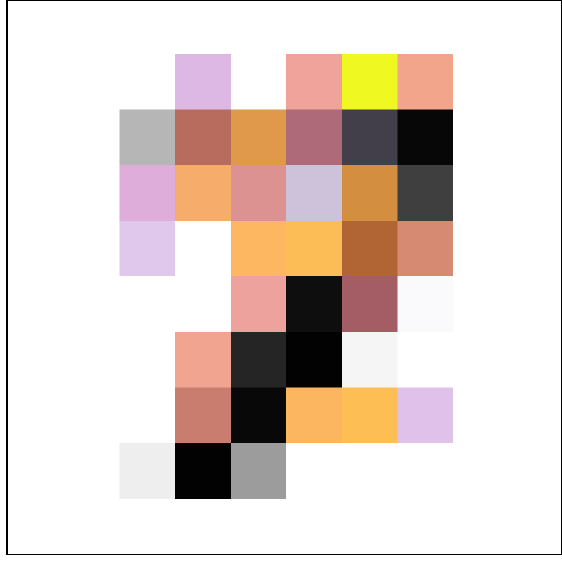}
    {\small \ffalabel{1.2k}}
    \label{fig:1010-1v7-1200s-wffa}
   \end{subfigure}%
   \hspace{0.01cm}
   \begin{subfigure}[b]{0.09\textwidth}
    \centering
    \includegraphics[width=\textwidth]{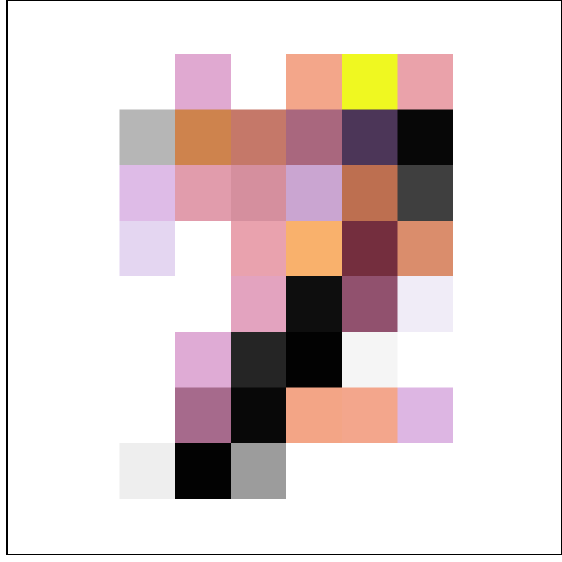}
    {\small FFA}
    \label{fig:1010-1v7-attr-wffa}
   \end{subfigure}%
     \caption{10 $\times$ 10 MNIST 1 vs. 7. The prediction is 7.}
    \label{fig:1010-1v7-wffa}
\end{figure}

\begin{figure}[t]
  \centering
   \begin{subfigure}[b]{0.09\textwidth}
    \centering
    \includegraphics[width=\textwidth]{10,10_pneumonia/10,10_pneumonia_inst28_ori}
    {\small Input}
    \label{fig:1010-pneumonia-input-wffa}
   \end{subfigure}%
   \hspace{0.01cm}
   \begin{subfigure}[b]{0.09\textwidth}
    \centering
    {\includegraphics[width=\textwidth]{10,10_pneumonia/10,10_pneumonia_inst28_lime_ori}}
    {\small LIME}
    \label{fig:1010-pneumonia-lime-wffa}
   \end{subfigure}%
   \hspace{0.01cm}
   \begin{subfigure}[b]{0.09\textwidth}
    \centering
    {\includegraphics[width=\textwidth]{10,10_pneumonia/10,10_pneumonia_inst28_shap_ori}}
    {\small SHAP}
    \label{fig:1010-pneumonia-shap-wffa}
   \end{subfigure}%
   \hspace{0.01cm}
   \begin{subfigure}[b]{0.09\textwidth}
    \centering
    \includegraphics[width=\textwidth]{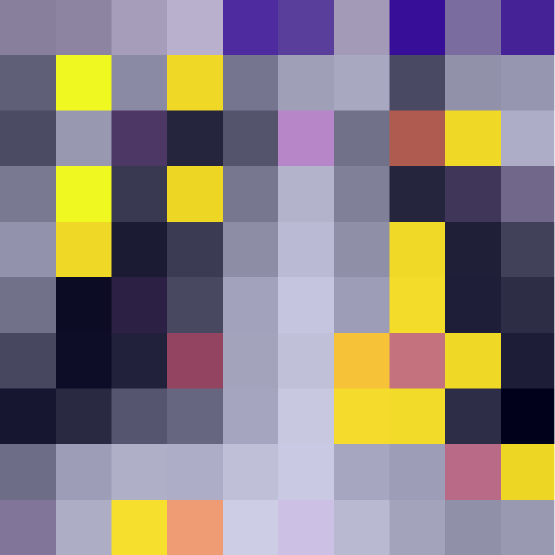}
    {\small \ffalabel{10}}
    \label{fig:1010-pneumonia-10s-wffa}
   \end{subfigure}%
   \hspace{0.01cm}
   \begin{subfigure}[b]{0.09\textwidth}
    \centering
    \includegraphics[width=\textwidth]{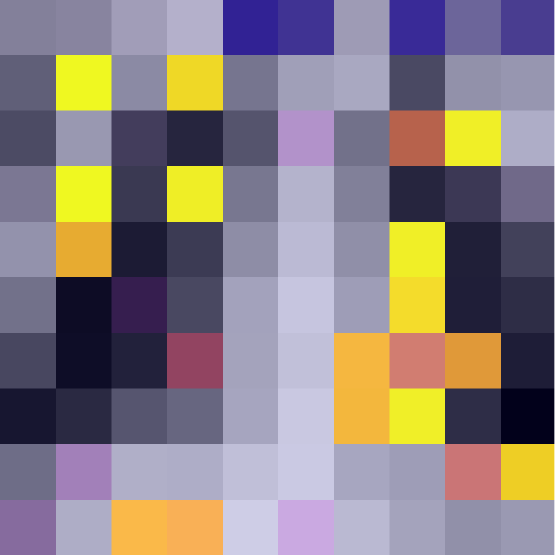}
    {\small \ffalabel{30}}
    \label{fig:1010-pneumonia-30s-wffa}
   \end{subfigure}%
   \hspace{0.01cm}
   \begin{subfigure}[b]{0.09\textwidth}
    \centering
    \includegraphics[width=\textwidth]{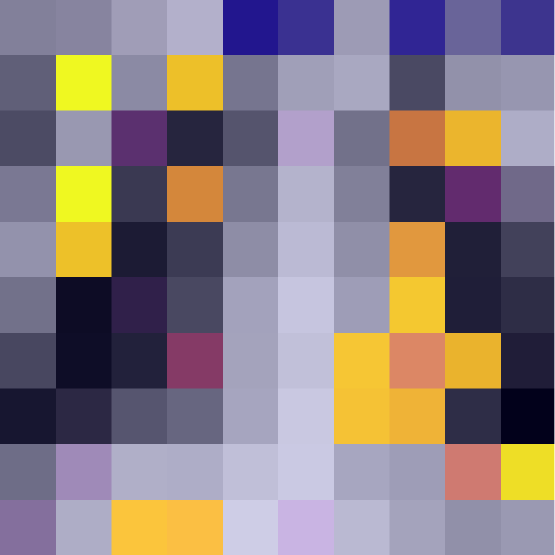}
    {\small \ffalabel{60}}
    \label{fig:1010-pneumonia-60s-wffa}
   \end{subfigure}%
   \hspace{0.01cm}
   \begin{subfigure}[b]{0.09\textwidth}
    \centering
    \includegraphics[width=\textwidth]{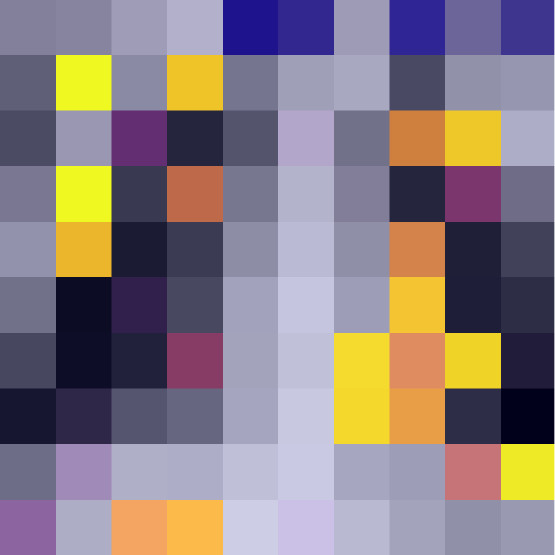}
    {\small \ffalabel{120}}
    \label{fig:1010-pneumonia-120s-wffa}
   \end{subfigure}%
   \hspace{0.01cm}
   \begin{subfigure}[b]{0.09\textwidth}
    \centering
    \includegraphics[width=\textwidth]{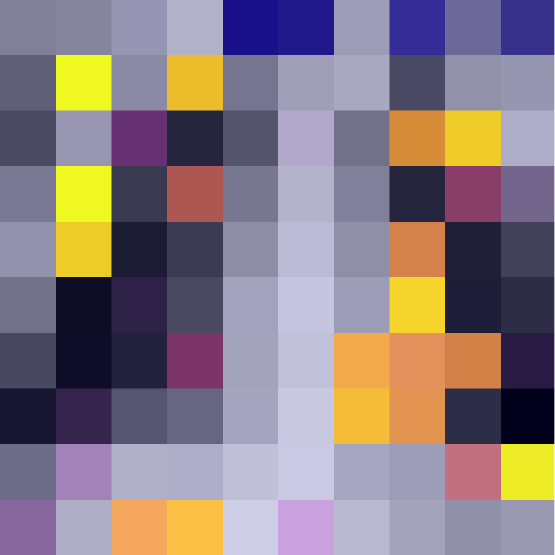}
    {\small \ffalabel{600}}
    \label{fig:1010-pneumonia-600s-wffa}
   \end{subfigure}%
   \hspace{0.01cm}
   \begin{subfigure}[b]{0.09\textwidth}
    \centering
    \includegraphics[width=\textwidth]{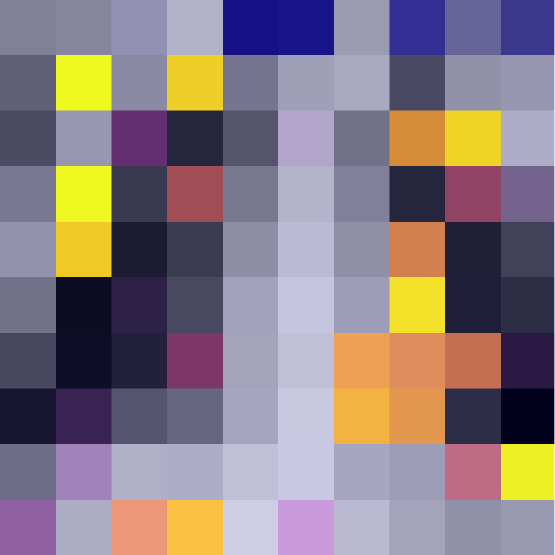}
    {\small \ffalabel{1.2k}}
    \label{fig:1010-pneumonia-1200s-wffa}
   \end{subfigure}%
   \hspace{0.01cm}
   \begin{subfigure}[b]{0.09\textwidth}
    \centering
    \includegraphics[width=\textwidth]{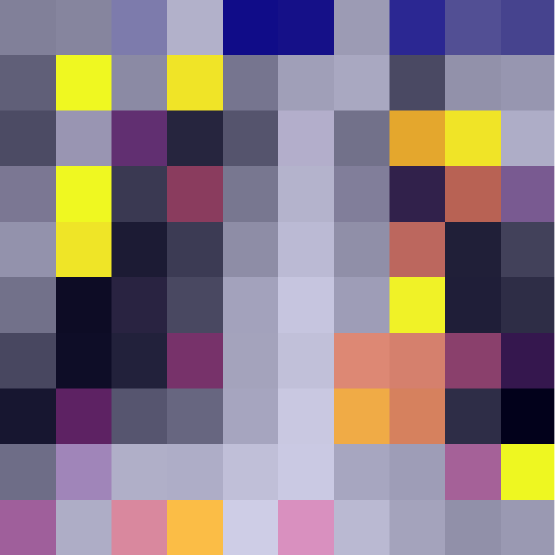}
    {\small FFA}
    \label{fig:1010-pneumonia-attr-wffa}
   \end{subfigure}%
     \caption{10 $\times$ 10 PneumoniaMNIST. The prediction is pneumonia.}
    \label{fig:1010-pneumonia-wffa}
\end{figure}

\autoref{tab:10wres} provides a comprehensive comparison of
approximate WFFA against feature attribution reported by LIME and SHAP
with respect to the exact WFFA values, conducted on the downscaled
MNIST digists and PneumoniaMNIST images, where exhaustive AXp
enumeration is feasible.
The values of feature attribution generated by LIME, SHAP, and
approximate~\wffalabel{$\ast$} for the three selected 10 $\times$ 10
images are shown in \autoref{fig:1010-1v3-wffa},
\autoref{fig:1010-1v7-wffa}, and \autoref{fig:1010-pneumonia-wffa}.
Over time, the number of features included in the AXp's increases, and
the weighted attribution of each feature changes converging to the
exact WFFA.
The results shown in \autoref{fig:1010-1v3}, \autoref{fig:1010-1v7},
and \autoref{fig:1010-pneumonia} align with the main finding for FFA
approximation shown earlier.
Furthermore, the results shown in \autoref{tab:10wres} are also
consistent with FFA observations in \autoref{tab:10res}.
%
%
Both LIME and SHAP can process each image within a runtime of less
than one second.
The average runtime and average number of AXp's generated for 10
$\times$ 10 MNIST 1 vs 3 (resp. 1 vs 7) are 14264.78s and 15781.87
(resp. 6834.61s and 4028.27), while the values in 10 $\times$ 10
PneumoniaMNIST are 8656.18s and 8802.87, respectively.
%
Similarly to the results in \autoref{tab:10res}, \autoref{tab:10wres}
indicates that our approximation yields small errors.
Even after 10 seconds, it outperforms both LIME and SHAP, and the
errors continue to decrease as we compute more AXp's.
Once again, the results of the orderings demonstrate that after 10
seconds, the ordering of \wffalabel{$\ast$} approaches closer to the
exact WFFA compared to both LIME and SHAP and converges to the exact
WFFA ordering with the growth of the number AXp's enumerated.
As can also be seen, LIME exhibits a substantial distance from the
\emph{exact} WFFA ordering.

\subsection{Summary}

The findings of this section again indicate that we can confidently
obtain valuable approximations of the exact WFFA values without the
need to exhaustively enumerate all AXp's for a given data instance.
It is worth noting that feature attribution determined by LIME and
SHAP is quite inaccurate and does not provide meaningful insights to a
human decision-maker, despite being computationally fast.

\begin{table}[t!]
\caption{LIME and SHAP versus WFFA on tabular data.}
\label{tab:tabwffa}
\centering
\scalebox{0.77}{
	\setlength{\tabcolsep}{1.2pt}
	\begin{tabular}{cccccccccccc}  \toprule
\textbf{Dataset} & \textbf{adult} & \textbf{appendicitis} & \textbf{australian} & \textbf{cars} & \textbf{compas} & \textbf{heart-statlog} & \textbf{hungarian} & \textbf{lending} & \textbf{liver-disorder} & \textbf{pima} & \textbf{recidivism}  \\
 $|\fml{F}|$  & (12) & (7) & (14) & (8) & (11) & (13) & (13) & (9) & (6) & (8) & (15)  \\ \midrule
 \textbf{Approach} & \multicolumn{11}{c}{\textbf{Error}} \\ \midrule
LIME & 4.32 & 2.06 & 4.96 & 1.48 & 3.26 & 4.40 & 4.43 & 1.37 & 2.37 & 2.63 & 4.66  \\
SHAP & 4.29 & 1.87 & 4.31 & 1.36 & 2.63 & 3.61 & 4.00 & 1.43 & 2.25 & 2.91 & 4.67  \\  \midrule
 & \multicolumn{11}{c}{\textbf{Kendall’s Tau}} \\ \midrule
LIME & 0.11 & 0.17 & 0.25 & -0.08 & -0.08 & 0.22 & 0.08 & -0.35 & -0.17 & 0.25 & 0.08  \\
SHAP & 0.07 & 0.23 & 0.31 & -0.07 & -0.07 & 0.22 & 0.26 & -0.38 & -0.16 & 0.15 & 0.16  \\  \midrule
 & \multicolumn{11}{c}{\textbf{RBO}} \\ \midrule
LIME & 0.53 & 0.65 & 0.48 & 0.64 & 0.56 & 0.56 & 0.40 & 0.59 & 0.65 & 0.69 & 0.38  \\
SHAP & 0.48 & 0.67 & 0.55 & 0.66 & 0.59 & 0.52 & 0.49 & 0.61 & 0.67 & 0.64 & 0.43  \\  \bottomrule
	\end{tabular}
}
\end{table}

\begin{table}[t!]
\caption{Comparison on 10 × 10 Images of WFFA versus LIME, SHAP and WFFA approximations.}
\label{tab:10wres}
\centering
\scalebox{0.84}{
	\setlength{\tabcolsep}{3pt}
	\begin{tabular}{ccccccccc}  \toprule
\textbf{Dataset} & \textbf{LIME} & \textbf{SHAP} & \textbf{\wffalabel{10}} & \textbf{\wffalabel{30}} & \textbf{\wffalabel{60}} & \textbf{\wffalabel{120}} & \textbf{\wffalabel{600}} & \textbf{\wffalabel{1200}}  \\ \cmidrule{2-9}
$|\fml{F}|=100$ & \multicolumn{8}{c}{\textbf{Error}}  \\ \midrule
10$\times$10-mnist-1vs3 & 11.28 & 9.81 & 5.52 & 5.12 & 4.83 & 4.50 & 3.32 & 2.61  \\
10$\times$10-mnist-1vs7 & 12.46 & 8.11 & 4.07 & 3.47 & 2.83 & 2.38 & 1.34 & 0.97  \\
10$\times$10-pneumoniamnist & 17.25 & 17.84 & 5.33 & 4.29 & 3.76 & 3.36 & 2.20 & 1.63  \\  \midrule
 & \multicolumn{8}{c}{\textbf{Kendall's Tau}}  \\ \midrule
10$\times$10-mnist-1vs3 & -0.14 & 0.48 & 0.53 & 0.60 & 0.64 & 0.67 & 0.75 & 0.81  \\
10$\times$10-mnist-1vs7 & -0.33 & 0.47 & 0.58 & 0.65 & 0.73 & 0.79 & 0.86 & 0.90  \\
10$\times$10-pneumoniamnist & -0.02 & 0.24 & 0.67 & 0.74 & 0.80 & 0.81 & 0.90 & 0.92  \\  \midrule
 & \multicolumn{8}{c}{\textbf{RBO}}  \\ \midrule
10$\times$10-mnist-1vs3 & 0.20 & 0.50 & 0.63 & 0.67 & 0.70 & 0.74 & 0.81 & 0.84  \\
10$\times$10-mnist-1vs7 & 0.19 & 0.58 & 0.73 & 0.77 & 0.81 & 0.86 & 0.90 & 0.91  \\
10$\times$10-pneumoniamnist & 0.21 & 0.37 & 0.63 & 0.70 & 0.74 & 0.77 & 0.82 & 0.87  \\  \bottomrule
	\end{tabular}
}
\end{table}

\begin{figure}[t]
  \centering
   \begin{subfigure}[b]{0.09\textwidth}
    \centering
    {\includegraphics[width=\textwidth]{10,10_mnist_1v3_inst7_ori}}
    {\small Image}
    \label{fig:1010-1v3-input}
   \end{subfigure}%
   \hspace{0.01cm}
   \begin{subfigure}[b]{0.09\textwidth}
    \centering
    {\includegraphics[width=\textwidth]{10,10_mnist_1v3_inst7_lime_ori}}
    {\small LIME}
    \label{fig:1010-1v3-lime}
   \end{subfigure}%
   \hspace{0.01cm}
   \begin{subfigure}[b]{0.09\textwidth}
    \centering
    {\includegraphics[width=\textwidth]{10,10_mnist_1v3_inst7_shap_ori}}
    {\small SHAP}
    \label{fig:1010-1v3-shap}
   \end{subfigure}%
   \hspace{0.01cm}
   \begin{subfigure}[b]{0.09\textwidth}
    \centering
    {\includegraphics[width=\textwidth]{10,10_mnist_1v3_inst7_time10_ori}}
    {\small \ffalabel{10}}
    \label{fig:1010-1v3-10s}
   \end{subfigure}%
   \hspace{0.01cm}
   \begin{subfigure}[b]{0.09\textwidth}
    \centering
    {\includegraphics[width=\textwidth]{10,10_mnist_1v3_inst7_time30_ori}}
    {\small \ffalabel{30}}
    \label{fig:1010-1v3-30s}
   \end{subfigure}%
   \hspace{0.01cm}
   \begin{subfigure}[b]{0.09\textwidth}
    \centering
    {\includegraphics[width=\textwidth]{10,10_mnist_1v3_inst7_time60_ori}}
    {\small \ffalabel{60}}
    \label{fig:1010-1v3-60s}
   \end{subfigure}%
   \hspace{0.01cm}
   \begin{subfigure}[b]{0.09\textwidth}
    \centering
    {\includegraphics[width=\textwidth]{10,10_mnist_1v3_inst7_time120_ori}}
    {\small \ffalabel{120}}
    \label{fig:1010-1v3-120s}
   \end{subfigure}%
   \hspace{0.01cm}
   \begin{subfigure}[b]{0.09\textwidth}
    \centering
    {\includegraphics[width=\textwidth]{10,10_mnist_1v3_inst7_time600_ori}}
    {\small \ffalabel{600}}
    \label{fig:1010-1v3-600s}
   \end{subfigure}%
   \hspace{0.01cm}
   \begin{subfigure}[b]{0.09\textwidth}
    \centering
    {\includegraphics[width=\textwidth]{10,10_mnist_1v3_inst7_time1200_ori}}
    {\small \ffalabel{1.2k}}
    \label{fig:1010-1v3-1200s}
   \end{subfigure}%
   \hspace{0.01cm}
   \begin{subfigure}[b]{0.09\textwidth}
    \centering
    {\includegraphics[width=\textwidth]{10,10_mnist_1v3_inst7_attr_ori}}
    {\small FFA}
    \label{fig:1010-1v3-attr}
   \end{subfigure}%
     \caption{10 $\times$ 10 MNIST 1 vs. 3. The prediction is 3.}
    \label{fig:1010-1v3}
\end{figure}

\begin{figure}[t]
  \centering
   \begin{subfigure}[b]{0.09\textwidth}
    \centering
    \includegraphics[width=\textwidth]{10,10_mnist_1v7_inst8_ori}
    {\small Image}
    \label{fig:1010-1v7-input}
   \end{subfigure}%
   \hspace{0.01cm}
    \begin{subfigure}[b]{0.09\textwidth}
    \centering
    {\includegraphics[width=\textwidth]{10,10_mnist_1v7_inst8_lime_ori}}
    {\small LIME}
    \label{fig:1010-1v7-lime}
   \end{subfigure}%
   \hspace{0.01cm}
   \begin{subfigure}[b]{0.09\textwidth}
    \centering
    {\includegraphics[width=\textwidth]{10,10_mnist_1v7_inst8_shap_ori}}
    {\small SHAP}
    \label{fig:1010-1v7-shap}
   \end{subfigure}%
   \hspace{0.01cm}
   \begin{subfigure}[b]{0.09\textwidth}
    \centering
    \includegraphics[width=\textwidth]{10,10_mnist_1v7_inst8_time10_ori}
    {\small \ffalabel{10}}
    \label{fig:1010-1v7-10s}
   \end{subfigure}%
   \hspace{0.01cm}
   \begin{subfigure}[b]{0.09\textwidth}
    \centering
    \includegraphics[width=\textwidth]{10,10_mnist_1v7_inst8_time30_ori}
    {\small \ffalabel{30}}
    \label{fig:1010-1v7-30s}
   \end{subfigure}%
   \hspace{0.01cm}
   \begin{subfigure}[b]{0.09\textwidth}
    \centering
    \includegraphics[width=\textwidth]{10,10_mnist_1v7_inst8_time60_ori}
    {\small \ffalabel{60}}
    \label{fig:1010-1v7-60s}
   \end{subfigure}%
   \hspace{0.01cm}
   \begin{subfigure}[b]{0.09\textwidth}
    \centering
    \includegraphics[width=\textwidth]{10,10_mnist_1v7_inst8_time120_ori}
    {\small \ffalabel{120}}
    \label{fig:1010-1v7-120s}
   \end{subfigure}%
   \hspace{0.01cm}
   \begin{subfigure}[b]{0.09\textwidth}
    \centering
    \includegraphics[width=\textwidth]{10,10_mnist_1v7_inst8_time600_ori}
    {\small \ffalabel{600}}
    \label{fig:1010-1v7-600s}
   \end{subfigure}%
   \hspace{0.01cm}
   \begin{subfigure}[b]{0.09\textwidth}
    \centering
    \includegraphics[width=\textwidth]{10,10_mnist_1v7_inst8_time1200_ori}
    {\small \ffalabel{1.2k}}
    \label{fig:1010-1v7-1200s}
   \end{subfigure}%
   \hspace{0.01cm}
   \begin{subfigure}[b]{0.09\textwidth}
    \centering
    \includegraphics[width=\textwidth]{10,10_mnist_1v7_inst8_attr_ori}
    {\small FFA}
    \label{fig:1010-1v7-attr}
   \end{subfigure}%
     \caption{10 $\times$ 10 MNIST 1 vs. 7. The prediction is 7.}
    \label{fig:1010-1v7}
\end{figure}

\begin{table}[t!]
\caption{Comparison on 28 × 28 Images of \wffalabel{7.2k} versus LIME, SHAP and WFFA approximations.}
\label{tab:28wres}
\centering
\scalebox{0.84}{
	\setlength{\tabcolsep}{3pt}
	\begin{tabular}{ccccccccc}  \toprule
\textbf{Dataset} & \textbf{LIME} & \textbf{SHAP} & \textbf{\wffalabel{10}} & \textbf{\wffalabel{30}} & \textbf{\wffalabel{120}} & \textbf{\wffalabel{600}} & \textbf{\wffalabel{1200}} & \textbf{\wffalabel{3600}}  \\ \cmidrule{2-9}
$|\fml{F}|=784$  & \multicolumn{8}{c}{\textbf{Error}}  \\ \midrule
28,28-mnist-1,3 & 49.28 & 22.33 & 9.22 & 7.50 & 6.69 & 4.50 & 3.08 & 2.75  \\
28,28-mnist-1,7 & 54.78 & 24.39 & 11.53 & 9.40 & 7.00 & 4.60 & 3.33 & 2.29  \\
28,28-pneumoniamnist & 62.88 & 31.46 & 8.17 & 7.74 & 5.67 & 4.85 & 3.75 & 3.08  \\  \midrule
 & \multicolumn{8}{c}{\textbf{Kendall’s Tau}}  \\ \midrule
28,28-mnist-1,3 & -0.80 & 0.42 & 0.49 & 0.64 & 0.70 & 0.81 & 0.86 & 0.88  \\
28,28-mnist-1,7 & -0.79 & 0.34 & 0.43 & 0.57 & 0.72 & 0.82 & 0.87 & 0.92  \\
28,28-pneumoniamnist & -0.66 & 0.24 & 0.37 & 0.57 & 0.69 & 0.76 & 0.81 & 0.88  \\  \midrule
 & \multicolumn{8}{c}{\textbf{RBO}}  \\ \midrule
28,28-mnist-1,3 & 0.03 & 0.40 & 0.45 & 0.54 & 0.63 & 0.78 & 0.84 & 0.89  \\
28,28-mnist-1,7 & 0.03 & 0.34 & 0.41 & 0.47 & 0.60 & 0.74 & 0.81 & 0.91  \\
28,28-pneumoniamnist & 0.03 & 0.23 & 0.30 & 0.35 & 0.43 & 0.59 & 0.65 & 0.81  \\  \bottomrule
	\end{tabular}
}
\end{table}

\begin{figure}[t]
  \centering
   \begin{subfigure}[b]{0.10\textwidth}
    \centering
    {\includegraphics[width=\textwidth]{28,28_mnist_1v3_inst13_lime_ori}}
    {\small LIME}
    \label{fig:2828-1v3-lime-wffa}
   \end{subfigure}%
   \hspace{0.01cm}
   \begin{subfigure}[b]{0.10\textwidth}
    \centering
    {\includegraphics[width=\textwidth]{28,28_mnist_1v3_inst13_shap_ori}}
    {\small SHAP}
    \label{fig:2828-1v3-shap-wffa}
   \end{subfigure}%
   \hspace{0.001cm}
   \begin{subfigure}[b]{0.10\textwidth}
    \centering
    {\includegraphics[width=\textwidth]{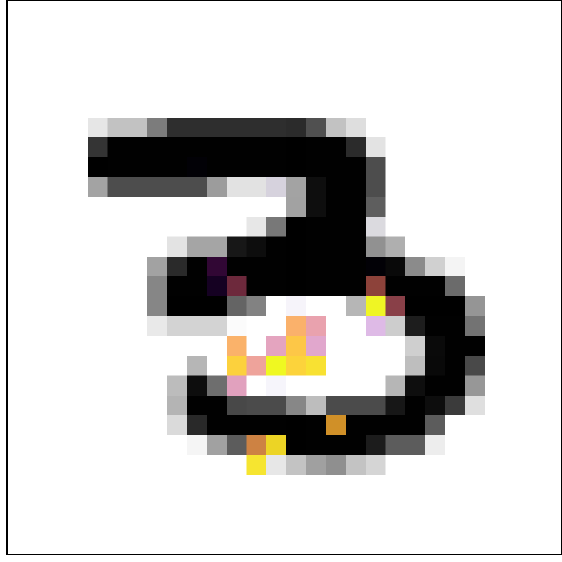}}
    {\small \wffalabel{10}}
    \label{fig:2828-1v3-10s-wffa}
   \end{subfigure}%
   \hspace{0.001cm}
   \begin{subfigure}[b]{0.10\textwidth}
    \centering
    {\includegraphics[width=\textwidth]{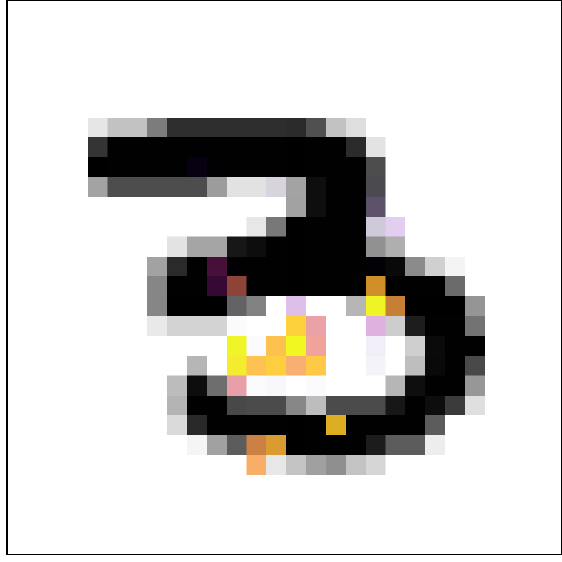}}
    {\small \wffalabel{30}}
    \label{fig:2828-1v3-30s-wffa}
   \end{subfigure}%
   \hspace{0.001cm}
   \begin{subfigure}[b]{0.10\textwidth}
    \centering
    {\includegraphics[width=\textwidth]{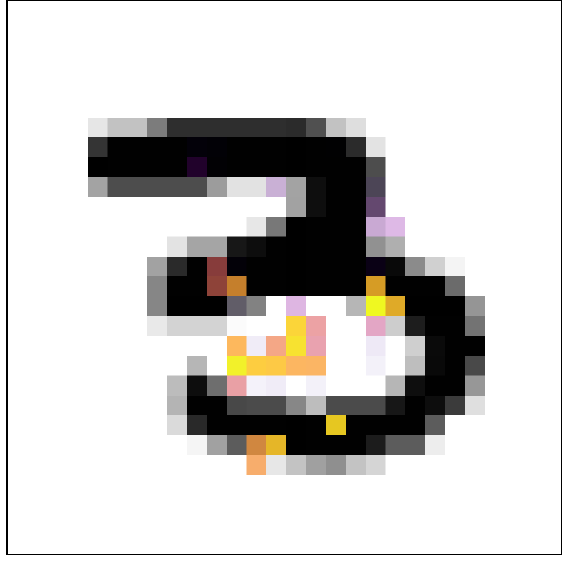}}
    {\small \wffalabel{120}}
    \label{fig:2828-1v3-120s-wffa}
   \end{subfigure}%
   \hspace{0.001cm}
   \begin{subfigure}[b]{0.10\textwidth}
    \centering
    {\includegraphics[width=\textwidth]{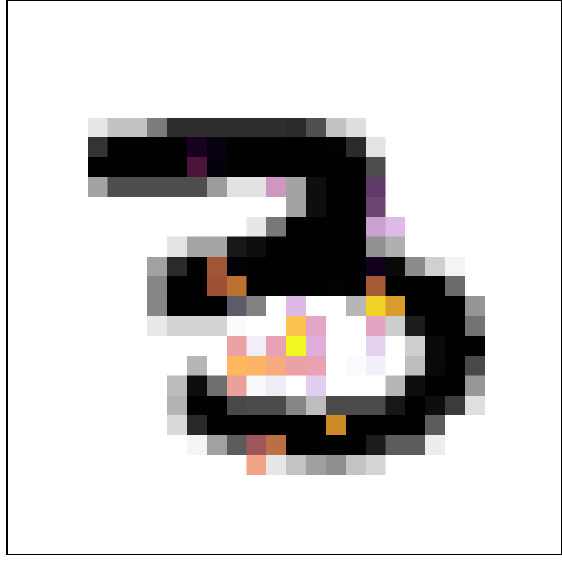}}
    {\small \wffalabel{600}}
    \label{fig:2828-1v3-600s-wffa}
   \end{subfigure}%
   \hspace{0.001cm}
   \begin{subfigure}[b]{0.10\textwidth}
    \centering
    {\includegraphics[width=\textwidth]{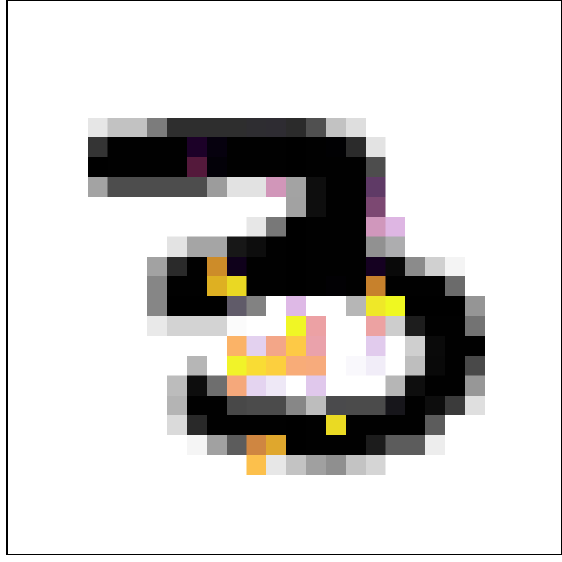}}
    {\small \wffalabel{1.2k}}
    \label{fig:2828-1v3-1200s-wffa}
   \end{subfigure}%
   \hspace{0.001cm}
   \begin{subfigure}[b]{0.10\textwidth}
    \centering
    {\includegraphics[width=\textwidth]{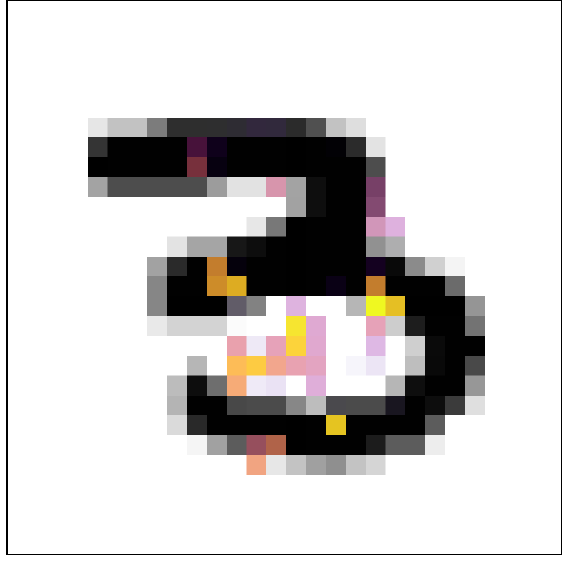}}
    {\small \wffalabel{3.6k}}
    \label{fig:2828-1v3-3600s-wffa}
   \end{subfigure}%
   \hspace{0.001cm}
   \begin{subfigure}[b]{0.10\textwidth}
    \centering
    {\includegraphics[width=\textwidth]{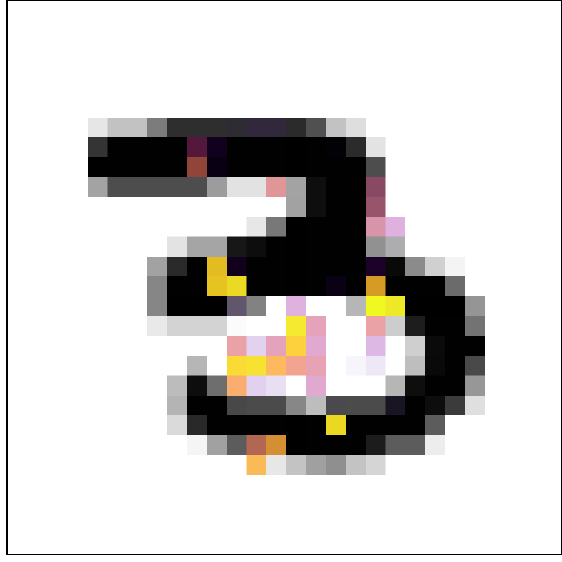}}
    {\small \wffalabel{7.2k}}
    \label{fig:2828-1v3-7200s-wffa}
   \end{subfigure}%
   \caption{28 $\times$ 28 MNIST 1 vs. 3. The prediction is digit 3.}
    \label{fig:2828-1v3-wffa}
\end{figure}

\begin{figure}[t]
  \centering
   \begin{subfigure}[b]{0.10\textwidth}
    \centering
    {\includegraphics[width=\textwidth]{28,28_mnist_1v7_inst4_lime_ori}}
    {\small LIME}
    \label{fig:2828-1v7-lime-wffa}
   \end{subfigure}%
   \hspace{0.01cm}
   \begin{subfigure}[b]{0.10\textwidth}
    \centering
    {\includegraphics[width=\textwidth]{28,28_mnist_1v7_inst4_shap_ori}}
    {\small SHAP}
    \label{fig:2828-1v7-shap-wffa}
   \end{subfigure}%
   \hspace{0.01cm}
   \begin{subfigure}[b]{0.10\textwidth}
    \centering
    {\includegraphics[width=\textwidth]{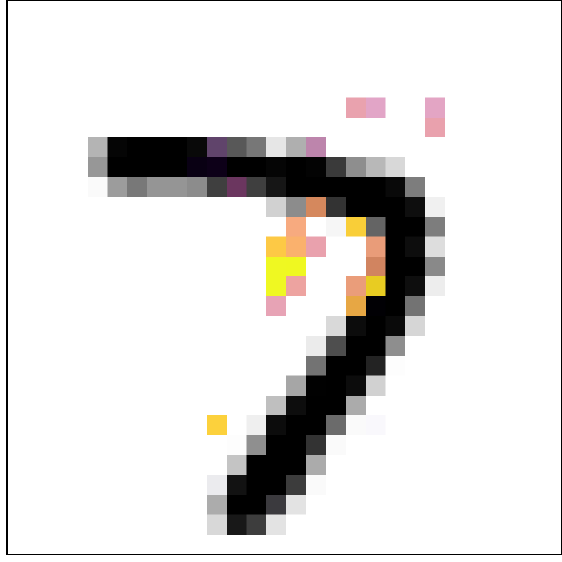}}
    {\small \wffalabel{10}}
    \label{fig:2828-1v7-10s-wffa}
   \end{subfigure}%
   \hspace{0.01cm}
   \begin{subfigure}[b]{0.10\textwidth}
    \centering
    {\includegraphics[width=\textwidth]{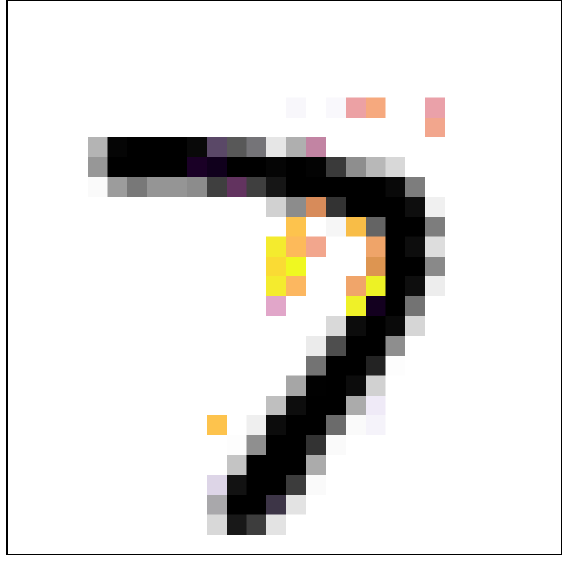}}
    {\small \wffalabel{30}}
    \label{fig:2828-1v7-30s-wffa}
   \end{subfigure}%
   \hspace{0.01cm}
   \begin{subfigure}[b]{0.10\textwidth}
    \centering
    {\includegraphics[width=\textwidth]{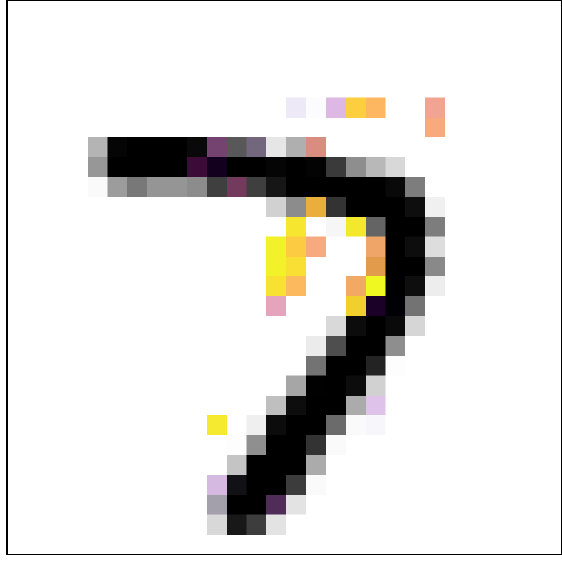}}
    {\small \wffalabel{120}}
    \label{fig:2828-1v7-120s-wffa}
   \end{subfigure}%
   \hspace{0.01cm}
   \begin{subfigure}[b]{0.10\textwidth}
    \centering
    {\includegraphics[width=\textwidth]{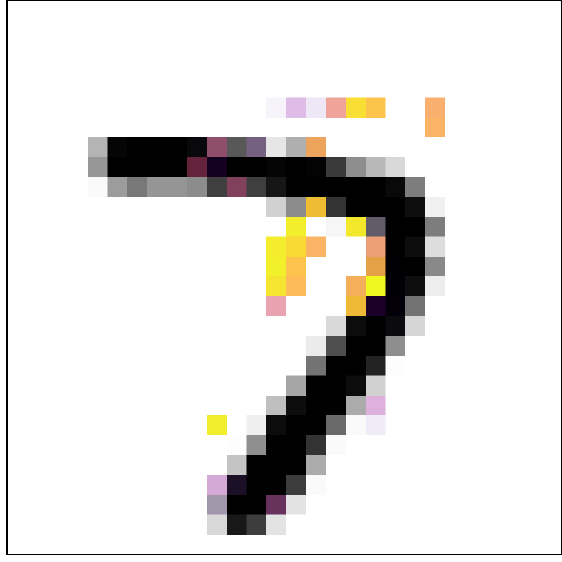}}
    {\small \wffalabel{600}}
    \label{fig:2828-1v7-600s-wffa}
   \end{subfigure}%
   \hspace{0.01cm}
   \begin{subfigure}[b]{0.10\textwidth}
    \centering
    {\includegraphics[width=\textwidth]{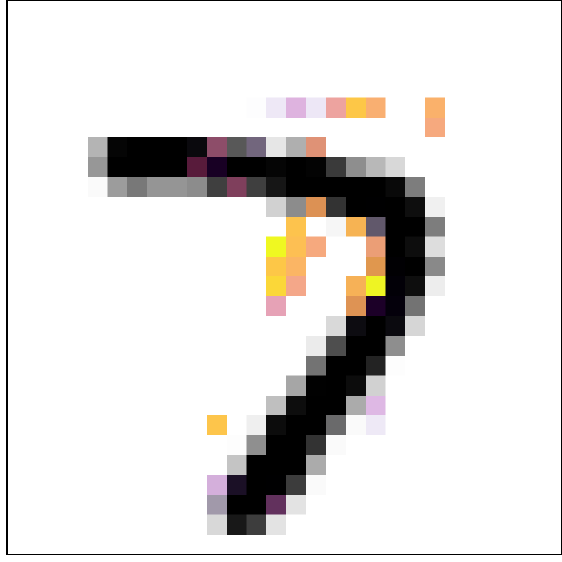}}
    {\small \wffalabel{1.2k}}
    \label{fig:2828-1v7-1200s-wffa}
   \end{subfigure}%
   \hspace{0.01cm}
   \begin{subfigure}[b]{0.10\textwidth}
    \centering
    {\includegraphics[width=\textwidth]{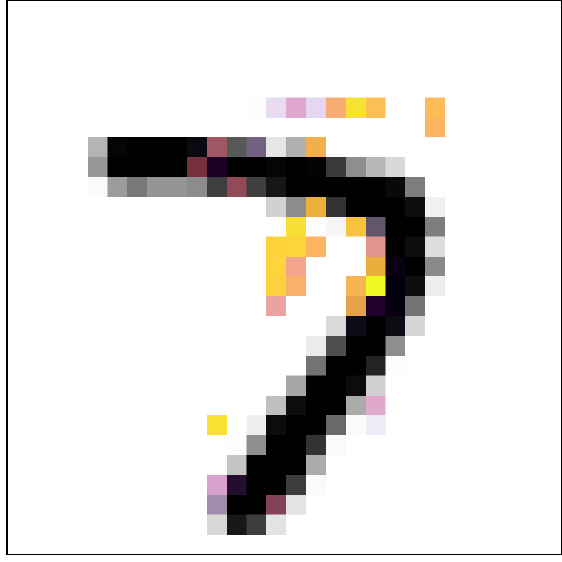}}
    {\small \wffalabel{3.6k}}
    \label{fig:2828-1v7-3600s-wffa}
   \end{subfigure}%
   \hspace{0.01cm}
   \begin{subfigure}[b]{0.10\textwidth}
    \centering
    {\includegraphics[width=\textwidth]{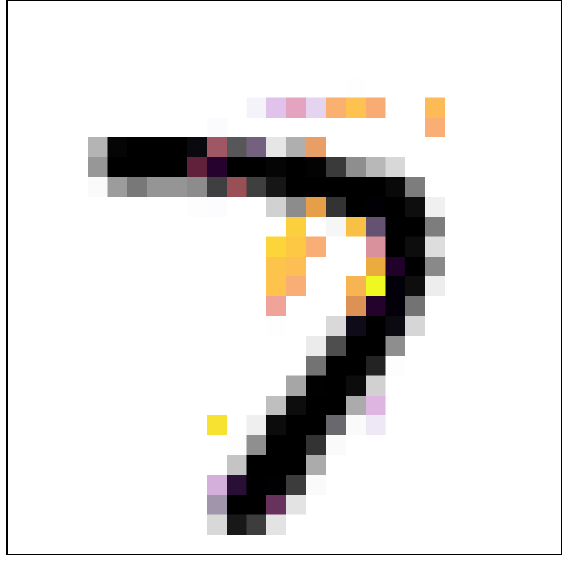}}
    {\small \wffalabel{7.2k}}
    \label{fig:2828-1v7-7200s-wffa}
   \end{subfigure}%
     \caption{28 $\times$ 28 MNIST 1 vs. 7. The prediction is digit 7.}
    \label{fig:2828-1v7-wffa}
\end{figure}

\begin{figure}[t]
  \centering
   \begin{subfigure}[b]{0.10\textwidth}
    \centering
    {\includegraphics[width=\textwidth]{28,28_pneumonia/28,28_pneumonia_inst9_lime_ori}}
    {\small LIME}
    \label{fig:2828-pneumonia-lime-wffa}
   \end{subfigure}%
   \hspace{0.01cm}
   \begin{subfigure}[b]{0.10\textwidth}
    \centering
    {\includegraphics[width=\textwidth]{28,28_pneumonia/28,28_pneumonia_inst9_shap_ori}}
    {\small SHAP}
    \label{fig:2828-pneumonia-shap-wffa}
   \end{subfigure}%
    \hspace{0.01cm}
   \begin{subfigure}[b]{0.10\textwidth}
    \centering
    {\includegraphics[width=\textwidth]{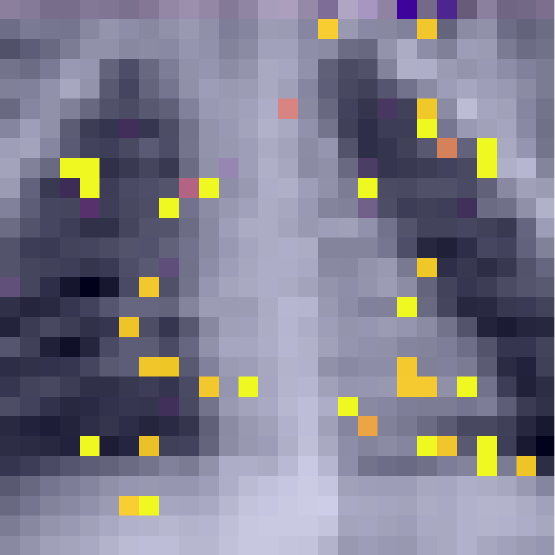}}
    {\small \wffalabel{10}}
    \label{fig:2828-pneumonia-10s-wffa}
   \end{subfigure}%
   \hspace{0.01cm}
   \begin{subfigure}[b]{0.10\textwidth}
    \centering
    {\includegraphics[width=\textwidth]{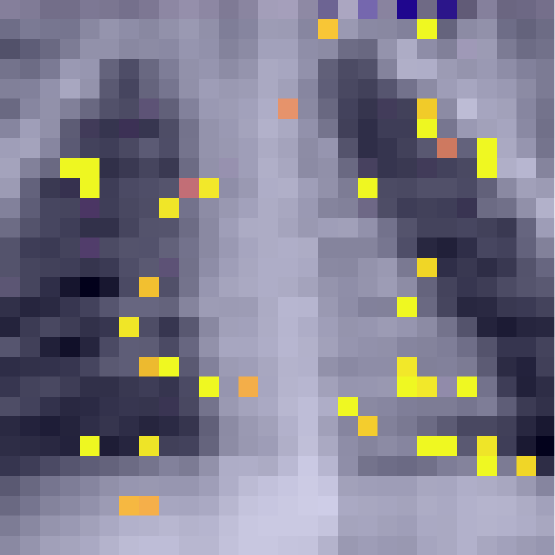}}
    {\small \wffalabel{30}}
    \label{fig:2828-pneumonia-30s-wffa}
   \end{subfigure}%
   \hspace{0.01cm}
   \begin{subfigure}[b]{0.10\textwidth}
    \centering
    {\includegraphics[width=\textwidth]{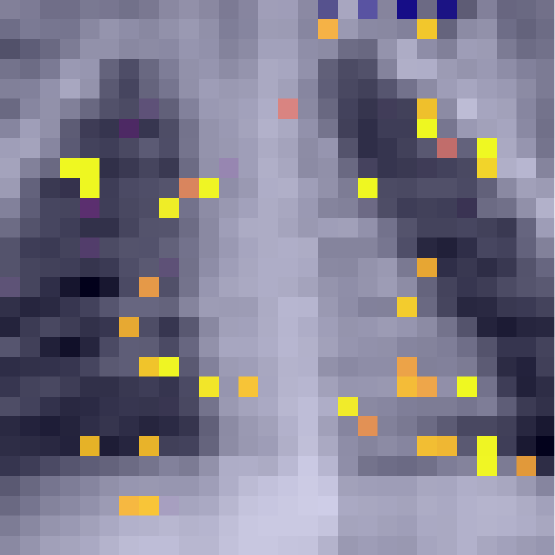}}
    {\small \wffalabel{120}}
    \label{fig:2828-pneumonia-120s-wffa}
   \end{subfigure}%
   \hspace{0.01cm}
   \begin{subfigure}[b]{0.10\textwidth}
    \centering
    {\includegraphics[width=\textwidth]{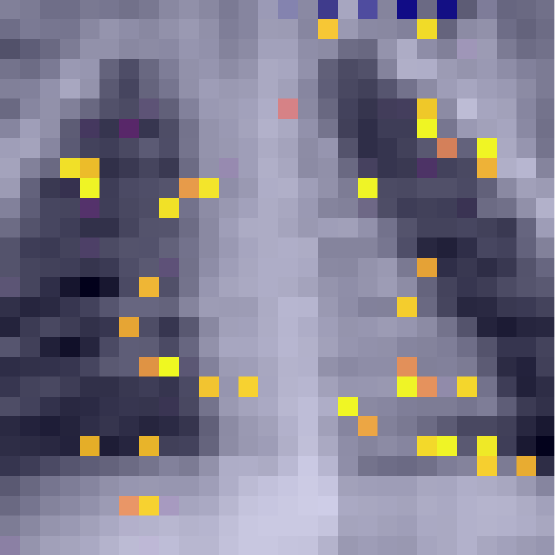}}
    {\small \wffalabel{600}}
    \label{fig:2828-pneumonia-600s-wffa}
   \end{subfigure}%
   \hspace{0.01cm}
   \begin{subfigure}[b]{0.10\textwidth}
    \centering
    {\includegraphics[width=\textwidth]{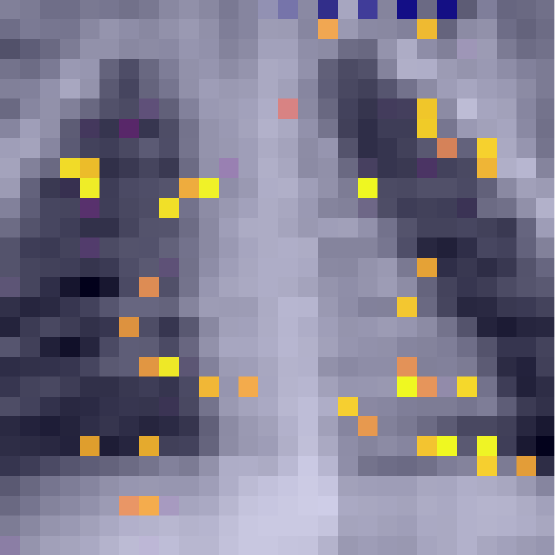}}
    {\small \wffalabel{1.2k}}
    \label{fig:2828-pneumonia-1200s-wffa}
   \end{subfigure}%
   \hspace{0.01cm}
   \begin{subfigure}[b]{0.10\textwidth}
    \centering
    {\includegraphics[width=\textwidth]{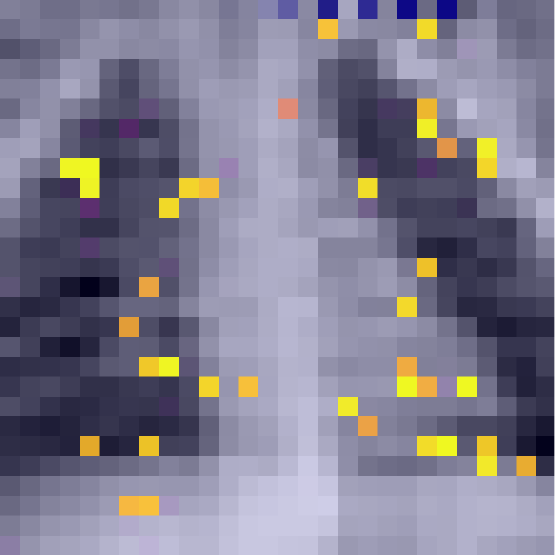}}
    {\small \wffalabel{3.6k}}
    \label{fig:2828-pneumonia-3600s-wffa}
   \end{subfigure}%
   \hspace{0.01cm}
   \begin{subfigure}[b]{0.10\textwidth}
    \centering
    {\includegraphics[width=\textwidth]{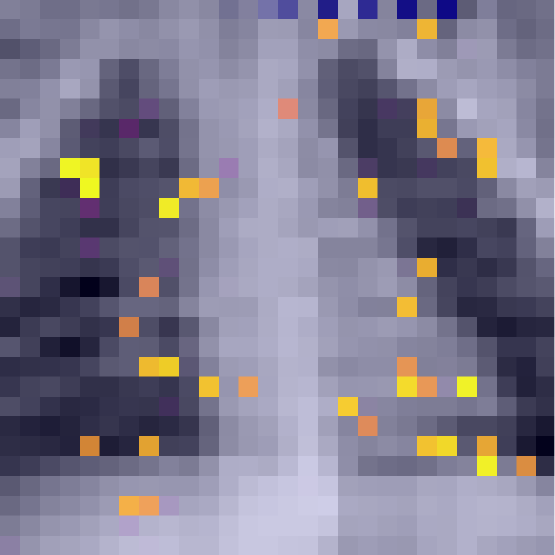}}
    {\small \wffalabel{7.2k}}
    \label{fig:2828-pneumonia-7200s-wffa}
   \end{subfigure}%
     \caption{28 $\times$ 28 PneumoniaMNIST. The prediction is normal.}
    \label{fig:2828-pneumonia-wffa}
\end{figure}

\begin{table}[t!]
\caption{Just-in-time Defect Prediction comparison of WFFA versus LIME and SHAP.}
\label{tab:jit-wffa}
\centering
\scalebox{0.83}{
	\begin{tabular}{ccccccc}  \toprule
	  \multirow{2}{*}{\textbf{Approach}} & \multicolumn{3}{c}{\textbf{openstack}~\small$(|\fml{F}|=\text{13})$} & \multicolumn{3}{c}{\textbf{qt}~\small$(|\fml{F}|= 16)$}  \\ \cmidrule{2-7}
  & \textbf{Error} & \textbf{kendalltau} & \textbf{rbo} & \textbf{Error} & \textbf{kendalltau} & \textbf{rbo}  \\ \midrule
LIME & 4.79 & 0.08 & 0.56 & 5.60 & -0.07 & 0.45  \\
SHAP & 5.01 & 0.02 & 0.54 & 5.17 & -0.11 & 0.44  \\  \bottomrule
	\end{tabular}
}
\end{table}

\section{Approximate Weighted Formal Feature Attribution}
%

As argued in \autoref{sec:fattr}, the exact WFFA computation can be
difficult in practice, due to the complexity of the problem.
But as \autoref{tab:10wres} indicates, our approach can yield decent
WFFA approximations even with a short duration of collecting AXp's.
Here we assess the fidelity of our approach in contrast to the
approximate WFFA computed after a duration of 2 hours~(7200s).
\wffalabel{$_\ast$} and the values of feature attribution generated by
LIME and SHAP for the three considered 28 $\times$ 28 images are
depicted in \autoref{fig:2828-1v3-wffa}, \ref{fig:2828-1v7-wffa}, and
\ref{fig:2828-pneumonia-wffa}.
As time progresses, the accumulated AXp's incorporate an increasing
number of features, and as a result the value of weighted attribution
for each feature can change.
\autoref{tab:28wres} details the comparison between LIME, SHAP, and
the approximate WFFA.
Both LIME and SHAP require less than one second to process each image.
The average results presented in \autoref{tab:28wres} are consistent
with those illustrated in \autoref{tab:10wres} and the FFA results
depicted in \autoref{tab:10res} and \autoref{tab:28res}.
\autoref{tab:28wres} demonstrates that after only 10 seconds, our WFFA
approximation outperforms both LIME and SHAP in terms of errors,
Kendall's Tau, and RBO values.
Additionally, after 10 seconds our approach produces weighted feature
attributions, which is closer to \wffalabel{7200} compared to both
LIME and SHAP.
This suggests that our approach effectively identifies the features
that are genuinely relevant for the prediction, which is in stark
contrast to LIME and SHAP.

\section{Application in Just-in-Time Defect Prediction}

Modern software companies often engage in the rapid and
frequent release of software products in short cycles.
Because of the exponential growth of
highly complex source code,
such rapid-release software development
presents significant challenges for
under-resourced Software Quality Assurance~(SQA) teams.
Developers are unable to thoroughly ensure the highest
quality of all newly developed code commits or pull
requests within the limited time and resources available,
due to the time-consuming and costly nature
of various SQA activities, e.g. code review.
To address this issue, a recent approach called Just-in-Time (JIT)
defect prediction~\cite{kim2007predicting,Kamei2013,pornprasit2021jitline,lin2021impact}
has been proposed.
This approach aims to predict whether a commit will introduce
software defects in the future such that development teams
can prioritize their limited SQA resources on the
riskiest commits or pull requests.

However, the JIT defect prediction approach
has frequently been criticized for being opaque
and lacking explainability for practitioners.
Model-agnostic explainability methods, e.g. LIME and SHAP,
cannot guarantee accurate feature attribution,
as discussed earlier in this appendix and
\autoref{sec:res}).
Experimental evidence presented in \autoref{sec:res} demonstrates the
usefulness of exact FFA in the context of JIT defect
prediction.
Given that our earlier observations above suggest that exact (resp.
approximate) WFFA is consistent with exact (resp. approximate) FFA, we
apply the computation of WFFA in the setting of JIT defection
prediction and demonstrate that it can be also a viable approach to
addressing practical explainability challenges.

In particular, where we use logistic regression models built on two
widely-used large-scale open-source datasets, namely Openstack and Qt,
which are commonly used in JIT defect prediction
studies~\cite{pornprasit2021pyexplainer}.
The property of monotonicity in logistic regression allows us to
enumerate explanations efficiently, following the approach
of~\cite{msgcin-icml21}.
By leveraging this method, we can extract the \emph{exact WFFA} for
each instance within one second.
The comparison of WFFA, LIME, and SHAP in terms of the three selected
metrics is provided in \autoref{tab:jit-wffa}.
These results are consistent with the FFA assessment presented in
\autoref{tab:jit}.
Similar to the findings in \autoref{tab:tabwffa},
\autoref{tab:10wres}, and \autoref{tab:28wres}, both LIME and SHAP
misalign with weighted formal feature attribution, although there are
some similarities between them.


\end{document}